\documentclass{article}

    \PassOptionsToPackage{numbers, compress}{natbib}

\usepackage[preprint]{neurips_2026}

\usepackage[utf8]{inputenc} 
\usepackage[T1]{fontenc}    
\usepackage{hyperref}       
\usepackage{url}            
\usepackage{booktabs}       
\usepackage{amsfonts}       
\usepackage{nicefrac}       
\usepackage{microtype}      
\usepackage{xcolor} 
\usepackage{graphicx}
\usepackage{amsmath}
\usepackage{hyperref}
\usepackage[nameinlink,capitalize,noabbrev]{cleveref}
\usepackage{wrapfig}
\usepackage{bbm}
\usepackage{multirow}
\usepackage[table]{xcolor}
\usepackage{tcolorbox}
\tcbuselibrary{skins,breakable}
\usepackage{array}
\usepackage{booktabs}
\usepackage{caption}
\usepackage{algorithm}

\usepackage{amsmath}
\usepackage{amssymb}
\usepackage{pifont}
\usepackage{algpseudocode}

\usepackage[nameinlink,capitalize,noabbrev]{cleveref}
\title{Reasoning Portability: Guiding Continual Learning for MLLMs in the RLVR Era}

\author{
Qiuhe Hong\textsuperscript{1},
Yuyang Liu\textsuperscript{1,}\thanks{Yuyang Liu(liuyuyang13@pku.edu.cn) and Yonghong Tian(yhtian@pku.edu.cn) are Corresponding Authors.} ,
Shuo Yang\textsuperscript{1},
Tiantian Peng\textsuperscript{1},
Fei Zhu\textsuperscript{2},
Yonghong Tian\textsuperscript{1,3,}\footnotemark[1]    
\\
\textsuperscript{1}Shenzhen Graduate School of Peking University
\\
\textsuperscript{2}Centre for Artificial Intelligence and Robotics, HKISI, CAS
\\
\textsuperscript{3}Peng Cheng Laboratory
}

\begin{document}

\maketitle

\begin{abstract}
  Vision-Language Models in Continual Learning (VLM-CL) aim to continuously adapt to new multimodal tasks while retaining prior knowledge. The emerging paradigm that couples Multimodal Large Language Models (MLLMs) with Reinforcement Learning with Verifiable Rewards (RLVR) calls for a new pattern to guide continual adaptation. Advances in reasoning capability now make it feasible to impose constraints at the reasoning level. We formalize \emph{portability}, a sample-level measure of how reusable the previous policy's behavior is on a new task, and empirically show that reasoning-level signals remain reliable on out-of-distribution samples while answer-level signals do not. We instantiate this as \emph{Reasoning Portability} (RP) and propose \textbf{R}easoning-based \textbf{D}ynamic \textbf{B}alance \textbf{C}ontinual \textbf{L}earning (\textbf{RDB-CL}), which modulates the per-sample Kullback-Leibler regularization in RLVR according to RP: a tight anchor preserves reusable reasoning on high-RP samples, while a relaxed anchor on low-RP samples permits exploration of new reasoning pathways. Experiments show that RDB-CL consistently outperforms baselines, improving \emph{Last} accuracy by +12.0\% over the vanilla RLVR baseline.
  Our code is available: \url{https://github.com/lluosi/RDB-CL}.


\end{abstract}

\section{Introduction}
\label{sec:intro}

Vision-Language Models Continual Learning (VLM-CL)~\citep{VLM-CLSurvey} aims to sequentially learn a stream of multimodal tasks while preventing the forgetting of previously learned knowledge. With the development of Large Language Models (LLMs)~\citep{llm} and  Multimodal Large Language Models (MLLMs)~\citep{mllm}, the focus of VLM-CL research has been gradually shifting from traditional lightweight architectures toward developing algorithms tailored for large-scale foundation models~\citep{hide,mllmcl}.

Traditional continual learning algorithms~\citep{clsurvey2, clsurvey} primarily  mitigate forgetting by applying regularization in either the parameter space (\eg EWC~\citep{ewc}, penalizing changes to important weights) or the feature space (\eg LwF~\citep{lwf}, using knowledge distillation to align representations). While effective for earlier architectures, these original approaches~\citep{ewc,lwf} are ill-suited for modern MLLMs in two complementary ways. First, the core capability of MLLMs is no longer representation extraction but emergent, high-level reasoning~\cite{deepseek}, which parameter-, 
feature-, or answer-constraints cannot directly govern. 
Second, the growing parameter scale and new post-training paradigm of MLLMs, \eg  Reinforcement Learning with Verifiable Rewards (RLVR)~\citep{tulu3,deepseekmath}, further hinder 
direct transplantation: representational regularizers have no natural interface with trajectory-level objectives.
Closing this gap calls for a reliable signal to guide continual learning under this new paradigm, one that captures whether the existing knowledge can transfer to new tasks.

\begin{figure}[t]
  \centering
  \includegraphics[height=4.7cm]{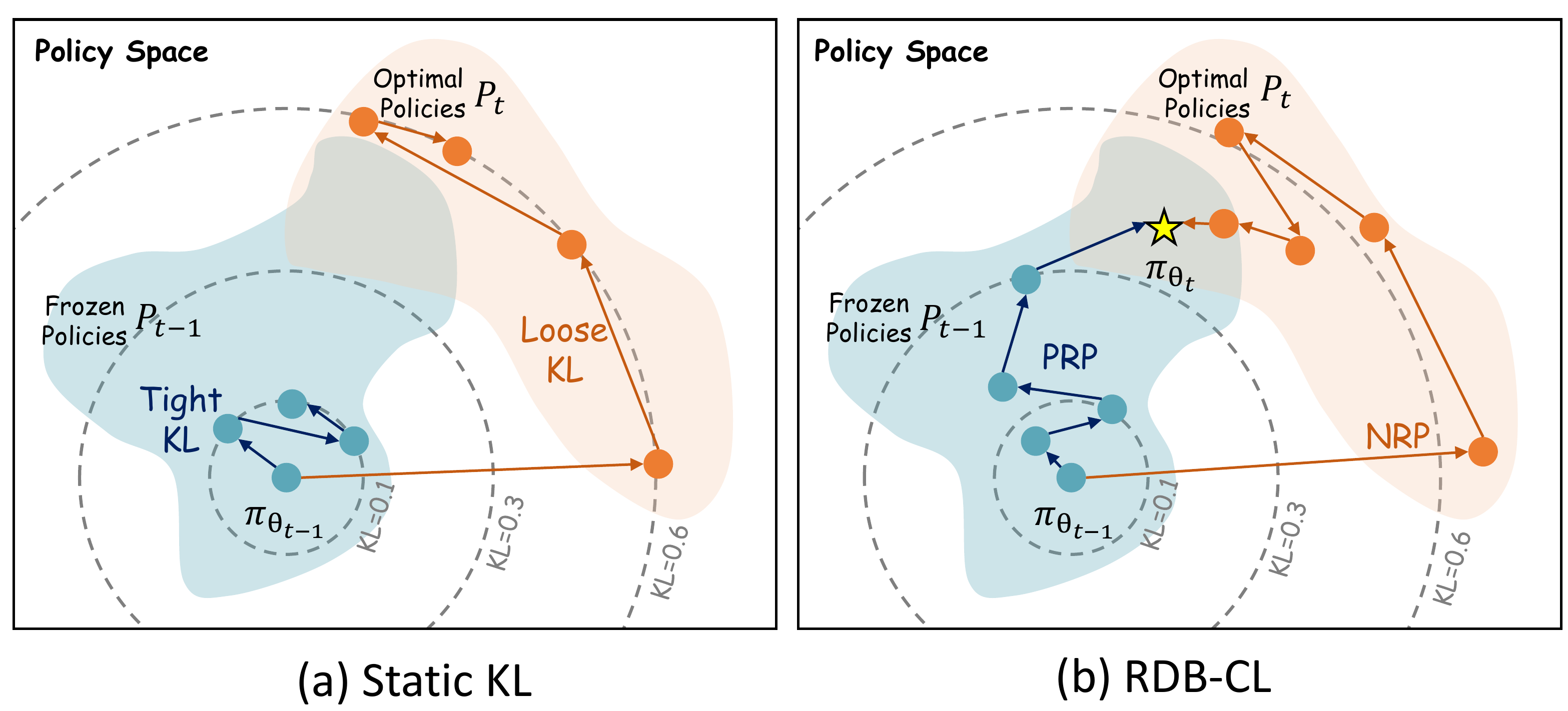}
  \caption{
  \textbf{Reasoning Portability guides per-sample adaptation in policy space.}
  (\emph{a}): A portability-agnostic constraint with static KL is uniform across samples, yielding either under-adaptation or forgetting.
  (\emph{b}): RDB-CL introduces reasoning portability (PRP: positive, NRP: negative) to modulate adaptation, steering updates toward the plasticity-stability trade-off.
  }
  \label{fig:illustration}
\end{figure}

We name this property \emph{portability}, which measures the extent to which the old policy's behavior remains reusable without revision. 
Recent advances in chain-of-thought reasoning with RLVR, such as Group-Relative Policy Optimization (GRPO)~\citep{deepseekmath}, make it feasible to directly optimize and interact with the reasoning process of MLLMs, rendering reasoning-level signals observable and actionable alongside answer-level ones.
Building on prior evidence that large models have the capability of self-evaluation~\citep{ptrue}, we compare answer- and reasoning-level confidence under distribution shift, treating each new-task sample as out-of-distribution (OOD) with respect to the previous policy. We find that while OOD samples hinder accurate prediction at the answer level, the reasoning itself remains consistent and transferable. 
We further investigate the stability and properties of this signal by case study.
These findings point toward reasoning-level guidance, and we instantiate portability at this level as \emph{Reasoning Portability} (RP), proxied by the previous policy's reasoning confidence.

With RP available as a per-sample signal, we propose 
\textbf{R}easoning-based \textbf{D}ynamic \textbf{B}alance 
\textbf{C}ontinual \textbf{L}earning (\textbf{RDB-CL}), a portability-aware continual RLVR framework for MLLMs. The KL divergence regularizer in RLVR serves as an anti-forgetting mechanism in the continual setting, but its commonly used static form imposes a rigid trade-off, sacrificing the plasticity needed to learn new tasks (\cref{fig:illustration}). Instead of applying a 
single constraint strength to every sample, RDB-CL stratifies new-task samples by their RP and modulates the per-sample KL coefficient accordingly: high-RP samples inherit a tight anchor to preserve reusable reasoning, while low-RP samples receive a relaxed anchor to permit exploration of new reasoning pathways. 
This design bridges the representational and reasoning levels, enabling continual adaptation without eroding the model’s procedural logic.

We evaluate RDB-CL on a visual question answering (VQA) continual learning benchmark across multiple task orders and obtain state-of-the-art results. A broad suite of studies, including ablations, hyperparameter sensitivity analyses, diversity preservation tests, and larger-scale experiments, demonstrates both robustness and the effectiveness of reasoning-guided updates.
In summary, our contributions are as follows:
\begin{itemize}
  \item We first identify reasoning-level confidence as a reliable signal for continual learning, better capturing cross-task transferability than answer-level. 
  \item We propose RDB-CL, an RLVR-based continual learning method that leverages \emph{reasoning portability} and adaptively modulates KL divergence regularization.
  \item We validate RDB-CL across diverse experiments, showing the enhanced plasticity-stability balance enabled by reasoning-level signals.
\end{itemize}
\section{Related Work}
\label{sec:related_work}

\paragraph{Post-training for MLLM with Reasoning.}

Post-training for MLLMs falls into supervised fine-tuning (SFT)~\citep{vllm, instructblip} and reinforcement learning (RL)~\citep{wang2024rlvlmfreinforcementlearningvision, zhai2024finetuninglargevisionlanguagemodels}. Compared with SFT, RL handles non-differentiable objectives, encourages exploration beyond supervised data, and offers stronger robustness against catastrophic forgetting~\citep{RL-MLLM-forgetting, RLRazor, retainingdoingroleonpolicy}. The paradigm has evolved from reinforcement learning from human feedback (RLHF)~\citep{ppo, dpo} to RLVR~\citep{tulu3} with verifiable objectives, while scalable optimizers such as GRPO and its variants~\citep{deepseekmath, dapo} further improve large-scale training. Following DeepSeek-R1~\citep{deepseekr1}, which showed that reasoning can self-emerge through GRPO without explicit supervision, recent works explore reasoning modules for MLLMs~\citep{vlrethinker, lookback}, ranging from modality-agnostic frameworks~\citep{r1_vl, mmeureka} to visual-specific chain-of-thought (CoT) mechanisms~\citep{mvot, deepeyes} that prove effective in compositional and complex visual reasoning. These advances make it feasible to incorporate reasoning modules into continual data streams, enabling MLLMs to adapt in a more stable and reasoning-consistent manner during post-training.

\paragraph{Continual Learning for MLLMs.}
Continual learning (CL) enables models to learn a task sequence without forgetting prior knowledge, while vision-language settings introduce additional failure modes such as cross-modal representation drift~\citep{VLM-CLSurvey}. Existing VLM-CL research predominantly builds on CLIP-style dual-encoder architectures, covering multi-modal replay~\citep{vqacl, mspt, incclip}, cross-modal regularization~\citep{dualteacher, scd, sgcl}, and parameter-efficient adaptation~\citep{proof, rail, diki}. However, these techniques are tightly coupled with contrastive dual-encoder representations and rarely transfer to modern MLLMs. Continual learning for MLLMs remains in its infancy, restricted to SFT under parameter-efficient fine-tuning (PEFT)~\citep{coin, hide}. Although continual reinforcement learning (CRL) algorithms have emerged for LLMs~\citep{cppo, mtcore}, RL-based CL for MLLMs remains unexplored. Moreover, most rely on expanding parameter spaces~\citep{mllmcl} or data pools~\citep{adaptinfty}, which is impractical under limited storage and raises governance risks when historical data are inaccessible or sensitive. These gaps highlight the need for an efficient CL framework tailored to modern MLLM fine-tuning, leveraging intrinsic signals between new and prior tasks for a better plasticity–stability trade-off.

\paragraph{Optimization via Confidence Scoring.}

Large models can develop meaningful self-evaluation of their own answer correctness~\citep{ptrue, internalconsistencyselffeedbacklargesurvey}, making confidence a reliable indicator for assessing outputs and mitigating hallucinations. Various estimators have emerged at sequence and semantic levels~\cite{selfconsistency, trustscore}, and confidence has been further incorporated into optimization as reward signals~\cite{huang2022largelanguagemodelsselfimprove}, data selection criteria~\cite{datauncertainty} or gradient weighting~\cite{prasad2025selfconsistencypreferenceoptimization}, with extensions to MLLMs~\cite{uncertaintyVLM}. With reasoning architectures, attention has shifted from answer-level to reasoning-level confidence with clearer benefits~\cite{reasoningconfidence, ucas}. In continual learning, reasoning-level confidence offers advantages over answer-level: it maintains a more stable, cross-task signal than task-specific answer confidence, which tends to vary with label spaces and formats.
\section{Finding a Reliable Signal for CL: Reasoning Portability}

\subsection{Portability in Continual Learning}

In continual learning, new-task samples are heterogeneous with respect to the prior policy, as some lie within the support of previously induced behaviors and others fall outside and require adaptation. We formalize this property as \emph{portability} $\mathcal{P}(x, \pi_{\theta_{t-1}}) \in [0,1]$: the degree to which prior behaviors remain reusable on a new-task sample $x \sim T_t$. Though rarely named, $\mathcal{P}$ is the latent objective behind many CL strategies that estimate inter-task relations in parameter, feature, or gradient space~\citep{lwf, gem}. We name it as the explicit target of estimation. This signal becomes especially relevant in large-scale post-training, where parameter-~\cite{ewc} and gradient-space~\cite{gem} estimators no longer scale, and adaptation must instead be steered through per-input, loss-level signals, for which $\mathcal{P}$ is a natural target as \cref{fig:illustration}. However, $\mathcal{P}$ is not directly observable during training, calling for a reliable \emph{observable proxy} to capture such property under distribution shift.

\begin{figure}[t]
  \centering
  \begin{minipage}[b]{0.46\linewidth}
  \centering
  \includegraphics[width=\linewidth]{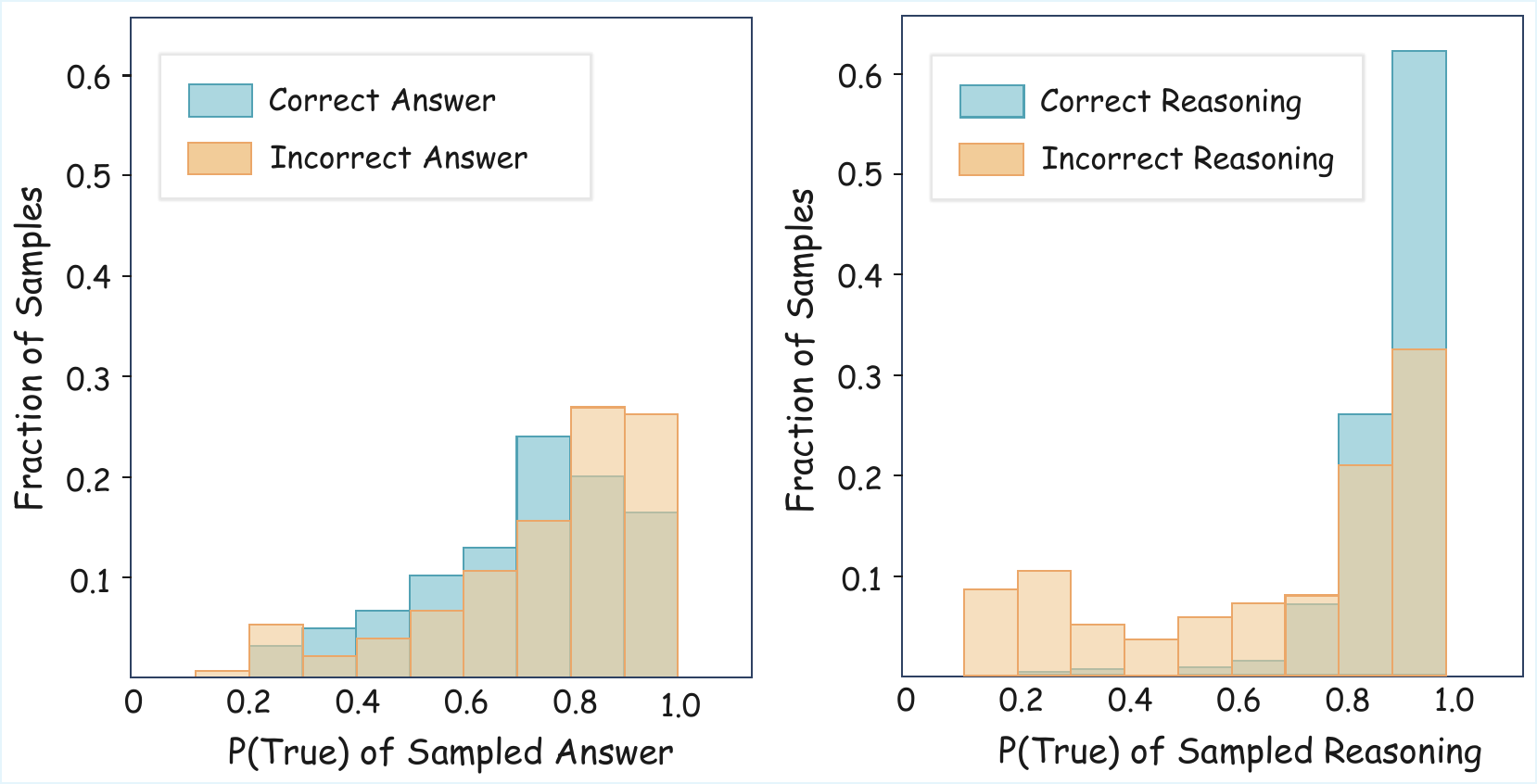}
  \caption{
  Confidence distribution for OOD tasks, measured as P(True). \emph{Left}: Distribution of answer confidence for correct vs. incorrect answers. \emph{Right}: Distribution of reasoning confidence for correct vs. incorrect reasoning, with reasoning labels provided by GPT-4o.
  }
 \label{fig:conf_dist}
\end{minipage}\hfill
  \begin{minipage}[b]{0.48\linewidth}
    \centering
    \includegraphics[width=\linewidth]{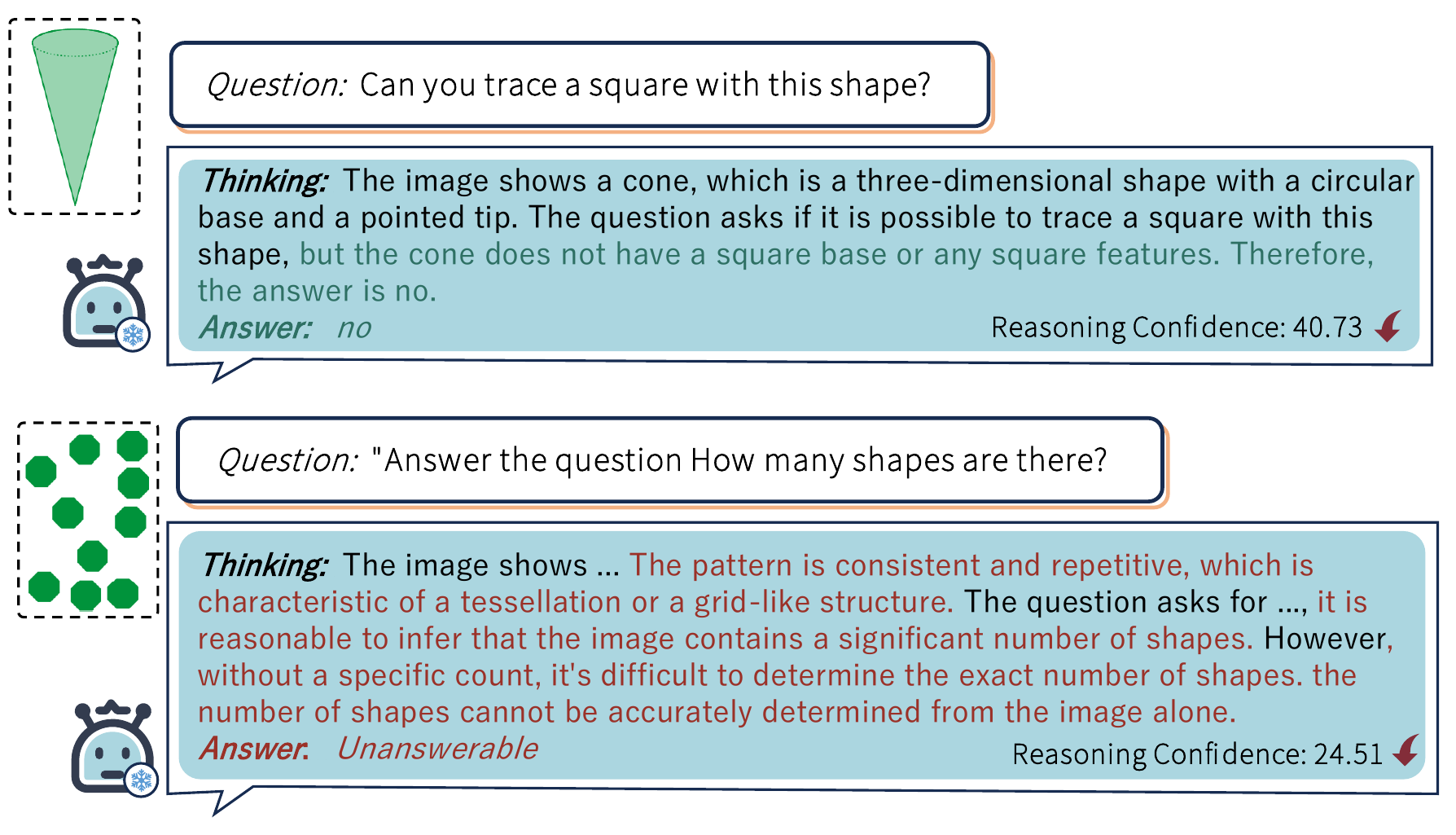}
    
    \caption{Low reasoning-confidence case study.
    IconQA examples after VizWiz$\rightarrow$ImageNet training show two typical sources of low confidence: \textcolor[HTML]{2E7064}{sound} reasoning under domain gap (\textit{Top}) and \textcolor[HTML]{9B3128}{incoherent} reasoning (\textit{Bottom}).
    }
    \label{fig:conf_cases}
  \end{minipage}
\end{figure}

\subsection{Empirical Portability Estimates under Distribution Shift}
\label{sebsec:confidence_study}

An intuitive proxy for $\mathcal{P}$ is the model's answer confidence, $C_{\theta}(a|x)$
which can evaluate whether the model itself knows this new knowledge~\citep{ptrue}. 
However, for OOD new-task samples, \cref{fig:conf_dist} (\emph{Left}) shows severe overlap between correct and incorrect answer distributions. This renders $C_{\theta}(a|x)$ a noisy, near-random guide that risks indiscriminate adaptation.

 Motivated by the reasoning capabilities of MLLMs, we hypothesize \emph{reasoning-level confidence} $C_{\theta}(r|x)$ as a more robust alternative. Unlike task-specific answers, the logical coherence of a trajectory $r$ transfers across domains. \cref{fig:conf_dist} (\emph{Right}) visually validates this: $C_{\theta}(r|x)$ exhibits clear separability between correct and incorrect reasoning. Most low-confidence samples correspond to incorrect reasoning, while high-confidence samples align with correct reasoning. Moreover, this separability persists across sequential tasks (Appendix \cref{fig:rc_degradation}), proving it is robust to task interference rather than an artifact of zero-shot ability.

\subsection{Reasoning Confidence as a Proxy for Reasoning Portability}
\label{subsec:confidence_case}

Given its stability, we instantiate $\mathcal{P}$ at the reasoning level as \emph{Reasoning Portability} (RP). For $x \in T_t$, RP dictates whether the previous policy's trajectory $r^{t-1} \sim \pi_{\theta_{t-1}}$ should be preserved. We formally adopt the reasoning confidence as its empirical proxy:
\begin{equation}
    \text{RP}(x) \approx C_{\theta_{t-1}}(r^{t-1}|x)
\end{equation}
Crucially, this proxy measures trajectory coherence, independent of final answer correctness.

To intuitively understand how $\text{RP}(x)$ captures the need for adaptation, \cref{fig:conf_cases} breaks down low-RP scenarios ($\text{RP}(x) \downarrow$) into two regimes: (i) \textbf{Domain-gap low confidence} (\textit{Top}): reasoning is sound, but severe visual shifts lower confidence; (ii) \textbf{Reasoning-failure low confidence} (\textit{Bottom}): the trajectory is internally incoherent. In both regimes, a low $\text{RP}(x)$ correctly signals that the previous policy's repertoire cannot confidently resolve $x$, effectively demarcating the boundary for required exploration. A parallel study on high-RP cases is in Appendix~\ref{subsec:more_confidence_cases}.

\section{RDB-CL: Portability-Aware Continual RLVR}
\label{sec:method}

\begin{figure}[tb]
  \centering
  \includegraphics[width=1\linewidth]{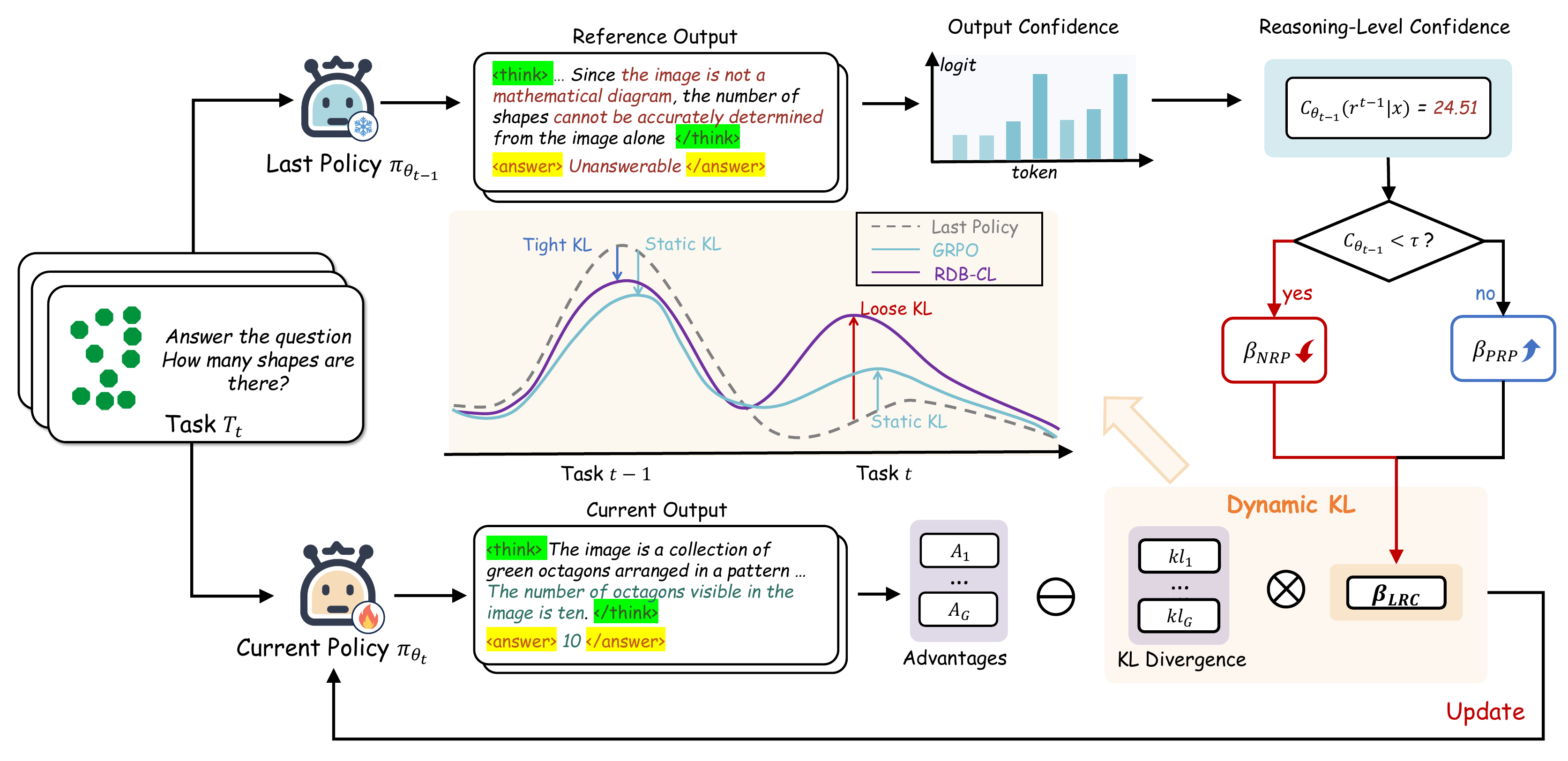}
  \caption{
  \textbf{Workflow of RDB-CL.} The previous policy $\pi_{\theta_{t-1}}$ generates reference responses for the current task $T_t$. Their $RP$ proxy $C_{\theta_{t-1}}(r^{t-1}|x)$ is used to classify samples as PRP or NRP by $\tau$. PRP  inherits the baseline KL anchor $\beta_0$ to preserve the previous policy's reasoning, while NRP receives a relaxed anchor $\beta_{LRC} < \beta_0$ to permit exploration of new reasoning pathways.
  }
  \label{fig:workflow}
\end{figure}

Building on this insight, we propose a \textbf{R}easoning-based \textbf{D}ynamic \textbf{B}alance \textbf{C}ontinual \textbf{L}earning (\textbf{RDB-CL}) method, which utilizes reasoning portability to guide continual adaptation in RLVR. Many methods rely on a KL penalty $D_{\mathrm{KL}}(\pi_\theta \,\|\, \pi_{ref})$ against a reference policy to stabilize training~\citep{ppo,deepseekmath}, which doubles as an implicit anti-forgetting mechanism when $\pi_{ref}$ is set to $\pi_{\theta_{t-1}}$ in the continual setting. However, they adopt a static coefficient $\beta$ that exhibits rigid behavior across different strengths (\cref{fig:feature_shift}), governing stability and plasticity uniformly. RDB-CL replaces this static constraint with a portability-gated one that varies per sample with reasoning portability, yielding a drop-in modification. We present the method under GRPO~\citep{deepseekmath}, a representative KL-regularized 
objective, and the overall workflow is shown in \cref{fig:workflow}.

\paragraph{Sample Stratification by RP.}  
To realize this gating, we partition new-task instances into two strata to drive distinct learning dynamics under the KL anchor. This follows advantage-based RL, where samples are split into positive and negative sets to induce distinct gradient behaviors.
Concretely, we threshold $RP$ score $\ C_{\theta_{t-1}}(r^{t-1}|x)$ by $\tau$:

\begin{equation}
    g_{\mathrm{RP}} =
    \begin{cases}
        \text{Negative (NRP)}, & if \ C_{\theta_{t-1}}(r^{t-1}|x) < \tau \\
        \text{Positive (PRP)}, &   if \  C_{\theta_{t-1}}(r^{t-1}|x) \ge \tau
    .
    \end{cases}
    \label{eq:rp_classify}
\end{equation}

\paragraph{Dynamic KL Modulation.} 
We modulate the KL penalty via a per-sample dynamic coefficient $\beta_{LRC}$ according to the $RP$ class, implementing a portability-gated relaxation of the constraint:
\begin{itemize}
    \item For Negative $RP$ (NRP) samples ($C < \tau$): The model must learn new reasoning pathways. We relax the KL constraint by scaling its weight proportionally to its (low) confidence $C_{\theta_{t-1}}(r^{t-1}|x)$, thereby promoting adaptation and plasticity.
    \item For Positive $RP$ (PRP) samples ($C \ge \tau$): The model should reuse existing mechanisms. We maintain a strong KL constraint (at its base value $\beta_0$) to maximize generalization and prevent feature drift near the old distribution.
\end{itemize}
To prevent degenerately small 
penalties, we impose a minimum clip $\mathit{clip}_{min}$. The dynamic per-sample KL coefficient $\beta_{LRC}$ is formally defined as:
\begin{equation}
    \beta_{LRC} = \max \left( clip_{min}, \mathbbm{1}_{\{C \ge \tau\}} + C \cdot \mathbbm{1}_{\{C < \tau\}} \right) \cdot \beta_0,
    \label{eq:beta_lrc}
\end{equation}
where $C_{\theta_{t-1}}(r^{t-1}|x)$, $\mathbbm{1}$ is the indicator function, and $clip_{min}$ is a minimum scaling factor to prevent degenerate penalties and maintain stability.

\paragraph{Final Objective Function.} Finally, we substitute this dynamic $\beta_{LRC}$ for the static $\beta$ in the standard GRPO objective to yield the final RDB-CL training objective:
\begin{equation}
    \begin{split}
        \mathcal{J}_{RDBCL}(\theta) = - \frac{1}{G} \sum_{i=1}^G \frac{1}{|o_i|} 
        \sum_{t=1}^{|o_i|} \left[ r_\theta(q,o_i,t) A_{i,t} \right. 
        \left. - \beta_{LRC} D_{\text{KL}} (\pi_{\theta_{t}} || \pi_{\theta_{t-1}}) \right].
    \end{split}
 \label{eq:rnloss}
\end{equation}

The training pipeline of RDB-CL is summarized in \cref{alg:all}. 
\begin{algorithm}[h]
\small
\caption{\textbf{R}easoning-based \textbf{D}ynamic \textbf{B}alance \textbf{C}ontinual \textbf{L}earning}
\label{alg:all}

\begin{algorithmic}[1]
\Require A MLLM $\pi_{0}$, task sequence $\mathcal{T}=\{T_1,\dots,T_t\}$
\State $\pi_{\theta_0} \gets \pi_0$

\For{$i = 1,\ldots,t$} \Comment{tasks arrive sequentially}
  \State $\pi_\theta \gets \pi_{\theta_{i-1}}$,\quad $\pi_{\mathrm{ref}} \gets \pi_{\theta_{i-1}}$  \Comment{initialize current and reference models from previous model}
  \Repeat
    \State Sample $x_i \sim T_i$ and $\{(r^{i-1}_{n}, a^{i-1}_{n})\}_{n=1}^N \sim \pi_{\theta_{i-1}}(\cdot \mid x_i)$
    \State Compute $C_{\theta_{i-1}}(r^{i-1}\mid x_i)$ and $\beta_{\mathrm{LRC}}$ via \cref{eq:beta_lrc}
    \State Sample $\{(r^{i}_{g}, a^{i}_{g})\}_{g=1}^G \sim \pi_{\theta}(\cdot \mid x_i)$
    \State Update $\theta$ by minimizing $\mathcal{J}$ in \cref{eq:rnloss}
  \Until{convergence on $T_i$}
  \State $\pi_{\theta_i} \gets \pi_\theta$ \Comment{store the model trained on task $i$}
\EndFor
\State \Return $\pi_{\theta_t}$
\end{algorithmic}

\end{algorithm}
\section{Experiments}
\label{sec:experiment}

\subsection{Experimental Setup}
\label{subsec:exp_setup}
\paragraph{Benchmark and Metrics.}
We construct a new benchmark by curating three VQA tasks from the popular MLLM continual learning benchmarks CoIN~\citep{coin} and UCIT~\citep{hide}, selecting datasets with pronounced domain gaps and low zero-shot performance: VizWiz~\citep{vizwiz}, ImageNet~\citep{imagenet}, and IconQA~\citep{iconqa}. To prioritize informative samples, we select 1.5k/2.7k/2.3k training examples based on zero-shot accuracy variance (commonly regarded as the most valuable for training), while retaining the original test sets (3.6k/5.0k/0.7k samples).
While the per-task training sizes appear modest, RLVR rolls out multiple trajectories per sample, making the effective training compute comparable to SFT over substantially larger datasets; the scale is also in line with prompt budgets commonly used in recent RLVR studies~\citep{perceptionr1}.
Following~\citep{zscl, moeadapters}, we adopt \textit{Average (Avg.)} and \textit{Last} as the primary metrics, measuring mean performance across all nodes during training and final performance after the full task sequence, respectively. We also report \textit{Backward Transfer (BwT)}~\citep{gem} as a standard forgetting indicator, and \textit{Finetune}~\citep{coin} (performance right after each task's training) in selected analyses.
We consider two task orders: \textbf{Order I}: VizWiz → ImageNet → IconQA; \textbf{Order II}: VizWiz → IconQA → ImageNet. All extended experiments use \textbf{Order I} unless otherwise specified.

\paragraph{Baselines.}

To evaluate the effectiveness of RDB-CL, we compare it with the following baselines:
(1) The zero-shot performance of the base model and the SFT, both in a continual learning setting and a joint multi-task setup, with the latter commonly regarded as the performance upper bound for SFT in CL. 
(2) Architecture-based modern methods specifically designed for continual learning of MLLMs, including MoELoRA~\citep{coin} and HiDe-LLaVA~\citep{hide}.
(3) GRPO-based methods, including the vanilla GRPO, which serves as our primary reference baseline. We further compare GRPO's variants combined with typical regularization-based CL approaches such as EWC~\citep{ewc} and LwF~\citep{lwf}, since our approach is also grounded in regularization techniques. For computational efficiency, we adopt a variant of these methods tailored for MLLMs.

\paragraph{Implementation Details.}

We adopt Qwen2.5-VL-7B-Instruct~\citep{qwen2.5-VL} as the base model within the CoIN~\citep{coin} continual learning framework. We implement GRPO following~\citep{tan2025reason}, retaining the default training hyperparameters. Rollouts are generated with vLLM~\citep{vllm}, sampling 8 rollouts per sample with a sampling temperature of 0.7. 
The original KL regularization weight is set to $\beta_0=0.15$, and we set $\tau = 0.7$ with a $\mathit{clip}_{min}=0.2$. We use P(True)~\citep{ptrue} to quantify the confidence score, which has been validated as highly effective in subsequent studies~\citep{rlhf, reasoningconfidence}. All experiments are conducted on a single node with $6 \times$ NVIDIA A6000 GPUs.

\subsection{Main Results}
\label{subsec:main_experiment}
\cref{tab:main_result} reports task-averaged performance on the MLLM-CL benchmark. We focus on GRPO-based variants, targeting single-model self-evolution without parameter expansion.
RDB-CL consistently outperforms vanilla GRPO, with relative gains of \textbf{+5.0\%} in \textit{Avg.} accuracy and \textbf{+12.0\%} in \textit{Last} accuracy, and \textbf{+33.4\%} in \textit{BwT} for \textbf{Average}.
While GRPO's uniform-regularization restricts adaptation, it indiscriminately entangles task subspaces, leading to substantial drift and aggravated forgetting.
EWC and LwF impose representation-level regularization that mitigates drift but lacks an explicit notion of transferability, thereby further limiting exploration, especially when inter-task feature distributions diverge. SFT shows strong plasticity via its off-policy, teacher-forced maximum likelihood estimation (MLE) objective, yet its aggressive probability reallocation amplifies catastrophic forgetting. 
Architecture-based methods (MoELoRA, HiDe-LLaVA) resist forgetting through hard isolation in expanded parameter spaces, but remain capacity-bounded by their isolated budgets, capping the achievable ceiling and underscoring the need for frameworks tailored to new paradigms.
In contrast, RDB-CL estimates reasoning portability to dynamically modulate regularization, decoupling adaptation from drift and steering updates toward a shared cross-task region. It achieves the best overall performance with competitive forgetting control, without extra parameters, memory, or inference cost and only $\sim$13.6\% additional training time (details in Appendix ~\ref{subsec:cost}). Furthermore, these efficiency advantages are not tied to GRPO: it generalizes seamlessly to other RLVR frameworks like DAPO~\citep{dapo}, delivering consistent gains (see Appendix \ref{app:more_results}, \cref{tab:dapo_result}).

\newcommand{\numplus}[2]{#1} 
\newcommand{\numplustop}[2]{#1} 

\newcommand{\cmark}{\ding{51}}
\newcommand{\xmark}{\ding{55}}

\renewcommand{\arraystretch}{1.15}
\begin{table*}[!t]
  \small
\caption{Overall performance on the MLLM-CL benchmark using Qwen2.5-VL-7B-Instruct as the base model. We evaluate two task orders (\textbf{Order I} / \textbf{Order II}) and report \textit{Average}($\uparrow$), \textit{Last}($\uparrow$) accuracies and \textit{BwT}($\uparrow$) across tasks, where $\star$ denotes the upper bound and $\Delta\Theta$ indicates the use of additional parameter space. We additionally report \textbf{Average} summary by averaging the results across the two orders. RDB-CL delivers competitive performance.}

 \label{tab:main_result}
 \centering
 \setlength{\tabcolsep}{4.9pt}
 \small
 \begin{tabular}{llc  ccc  ccc  ccc}
 \toprule
 \multicolumn{2}{c}{\multirow{2}{*}{\textbf{Method}}}
 & \multirow{2}{*}{\textbf{$\Delta\Theta$}}
 & \multicolumn{3}{c}{\textbf{Order I}}
 & \multicolumn{3}{c}{\textbf{Order II}}
 & \multicolumn{3}{c}{\textbf{Average}} \\
 \cmidrule(lr){4-6} \cmidrule(lr){7-9} \cmidrule(lr){10-12}
 \multicolumn{2}{c}{}
 & 
 & \textit{Avg.}  & \textit{Last} & \textit{BwT} 
 & \textit{Avg.}  & \textit{Last}  & \textit{BwT} 
 & \textit{Avg.}  & \textit{Last}  & \textit{BwT} \\
 \midrule

 \multicolumn{2}{c}{Zero-shot}
 & \xmark
 & 33.16 & 33.16 & --
 & 33.16 & 33.16 & --
 & 33.16 & 33.16 & -- \\

 \multicolumn{2}{c}{SFT (Multi-task)$^{\star}$}
 & \xmark
 & 54.95 & 54.95 & --
 & 54.95 & 54.95 & --
 & 54.95 & 54.95 & -- \\

 \multicolumn{2}{c}{SFT (In order)}
 & \xmark
 & 42.84 & 49.75 & -21.97
 & 35.42 & 51.32 & -16.51
 & 39.13 & 50.54 & -19.24 \\

 \multicolumn{2}{c}{MoELoRA}
 & \cmark
 & 35.14 & 40.81 & -0.64
 & 38.09 & 41.40 & -0.22
 & 36.62 & 41.11 & -0.43 \\

 \multicolumn{2}{c}{Hide-LLaVA}
 & \cmark
 & 35.74 & 42.17 & 0.55
 & 36.98 & 41.98 & 0.59
 & 36.36 & 42.08 & 0.57 \\

 \midrule 
 \multicolumn{12}{l}{\textit{GRPO-based methods}} \\
 \midrule

 & GRPO
 & \xmark
 & \numplus{41.32}{(+0.0\%)} & \numplus{45.40}{(+0.0\%)} & \numplus{-7.84}{(+0.0\%)}
 & \numplus{41.37}{(+0.0\%)} & \numplus{48.86}{(+0.0\%)} & \numplus{-8.63}{(+0.0\%)}
 & \numplus{41.35}{(+0.0\%)} & \numplus{47.13}{(+0.0\%)} & \numplus{-8.24}{(+0.0\%)} \\

 & GRPO+EWC
 & \xmark
 & \numplus{41.94}{(+1.5\%)} & \numplus{51.49}{(+13.4\%)} & \numplus{-5.00}{(+36.2\%)}
 & \numplus{41.15}{(-0.5\%)} & \numplus{\underline{51.03}}{(+4.4\%)} & \numplus{-7.86}{(+8.9\%)}
 & 41.55 & \underline{51.26} & -6.43 \\

 & GRPO+LwF
 & \xmark
 & \numplus{\underline{42.49}}{(+2.8\%)} & \numplus{\underline{53.03}}{(+16.8\%)} & \numplustop{\textbf{-1.33}}{(+83.0\%)}
 & \numplus{\underline{42.21}}{(+2.0\%)} & \numplus{47.81}{(-2.1\%)} & \numplustop{\textbf{-4.69}}{(+45.7\%)}
 & \underline{42.35} & 50.42 & \textbf{-3.01} \\

 \rowcolor{gray!15}
 & \textbf{RDB-CL (Ours)}
 & \xmark
 & \numplustop{\textbf{44.49}}{(+7.8\%)} & \numplustop{\textbf{53.79}}{(+18.5\%)} & \numplus{\underline{-4.83}}{(+38.4\%)}
 & \numplustop{\textbf{42.33}}{(+2.3\%)} & \numplustop{\textbf{51.76}}{(+5.9\%)} & \numplus{\underline{-6.14}}{(+28.9\%)}
 & \numplustop{\textbf{43.41}}{(+5.0\%)} & \numplustop{\textbf{52.78}}{(+12.0\%)} & \numplus{\underline{-5.49}}{(+33.4\%)} \\
 \bottomrule
 \end{tabular}
\end{table*}

\renewcommand{\arraystretch}{1.0}
\vspace{-4pt}
\begin{figure}[t]
  \centering
  \begin{minipage}[t]{0.49\linewidth}
    \centering
    \includegraphics[width=\linewidth]{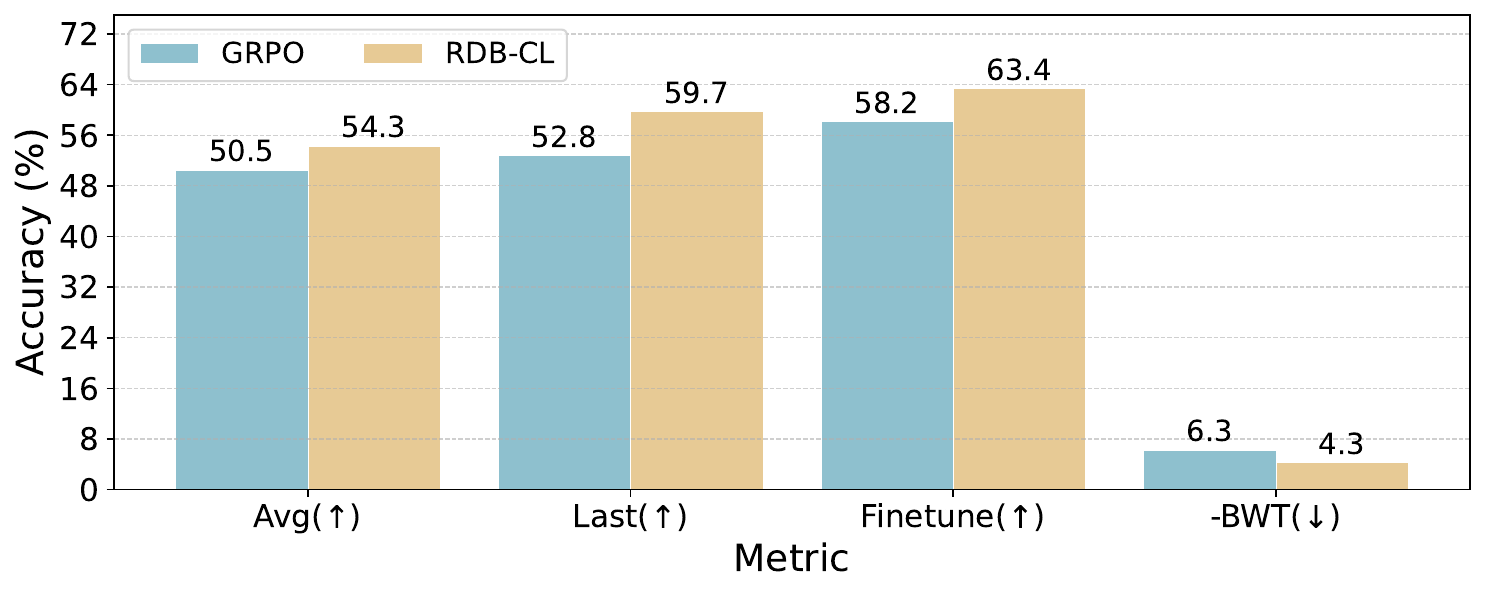}
    \caption{Results on a long-horizon benchmark.}
    \label{fig:longer_bars}
  \end{minipage}\hfill
  \begin{minipage}[t]{0.49\linewidth}
    \centering
    \includegraphics[width=\linewidth]{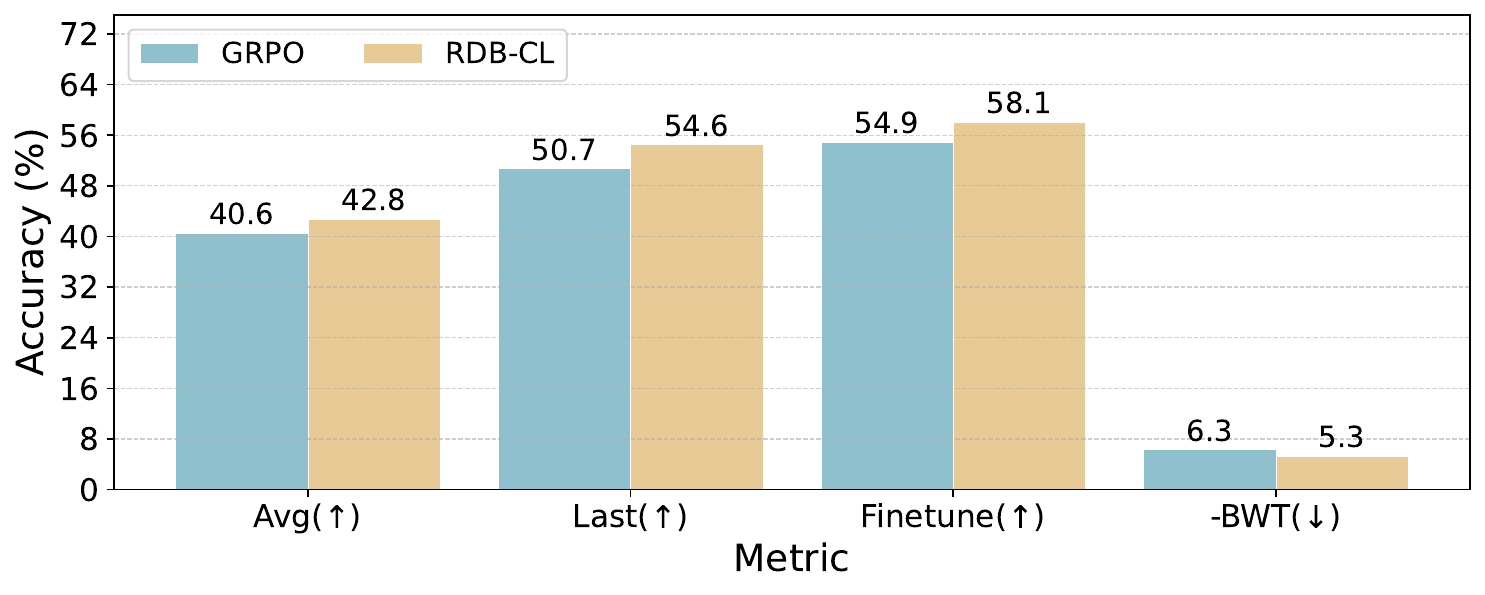}
    \caption{Results with larger per-task samples.}
    \label{fig:larger_bar}
  \end{minipage}
\end{figure}

\paragraph{Order Robustness.} 
We assess dependence on task order under \textbf{Order I} / \textbf{Order II}. As shown in \cref{tab:main_result}, RDB-CL achieves similar performance and consistent gains under both orders, indicating that the improvement is not tied to a particular task sequence. This robustness aligns with our design: reasoning confidence provides a reliable, sample-level estimate of portability under shift and steering adaptive updates largely independent of task ordering.

\subsubsection{Larger-Scale Benchmark Evaluation}
Due to the high training cost of full-parameter GRPO for MLLMs, our main experiments focus on three datasets and use downsampled training splits to enable controlled ablations. We then evaluate RDB-CL in larger-scale settings to test robustness under longer horizons and increased data regimes. The results align with the main setting, showing stable gains without exacerbating forgetting.

\paragraph{Longer benchmark.}

We extend the sequence to seven by adding TextVQA~\citep{textvqa}, ScienceQA~\citep{scienceqa}, VQAv2~\citep{vqav2} and GQA~\citep{gqa} from CoIN~\citep{coin}, expanding domain/answer coverage and stressing long-horizon error accumulation. As shown in \cref{fig:longer_bars}, RDB-CL preserves its gains over GRPO (+13.2\% \textit{Last}) without exacerbating forgetting (+31.8\% \textit{BwT}), confirming long-term stability.

\paragraph{Larger per-task samples.}
We scale per-task training data by $3\times$ over the main setting. As shown in \cref{fig:larger_bar}, the improvements of RDB-CL over GRPO persist under this increased data regime(+5.8\% \textit{Last} \& +15.9\% \textit{BwT}), suggesting low sensitivity to a particular sampling budget within this range. 

\subsection{Ablation for Core Claims Validation}

\begin{figure}[t]
  \centering
  \begin{minipage}[b]{0.46\linewidth}
    \centering
    \small
    \captionof{table}{Ablation results with dynamic KL modulated by different confidence signals.}
    \label{tab:answer_result}
    {\renewcommand{\arraystretch}{1.15}
\begin{tabular}{l|cccc}
\toprule
Method   & \textit{Avg.}  & \textit{Last} & \textit{Finetune} & \textit{BwT}\\ 
\midrule
Base Model    & 38.25  & 38.25  & 38.25 & -- \\
GRPO          & 41.32  & 45.40  & 50.63 & -7.84\\
w/ Answer     & 42.85  & 50.80  & 56.59 & -8.68\\
\rowcolor{gray!10}  
w/ Reasoning  & \textbf{44.49}  & \textbf{53.79}  & \textbf{57.01} & \textbf{-4.98}\\
\bottomrule
\end{tabular}
}
  \end{minipage}\hfill
  \begin{minipage}[b]{0.48\linewidth}
    \centering
    \includegraphics[width=\linewidth]{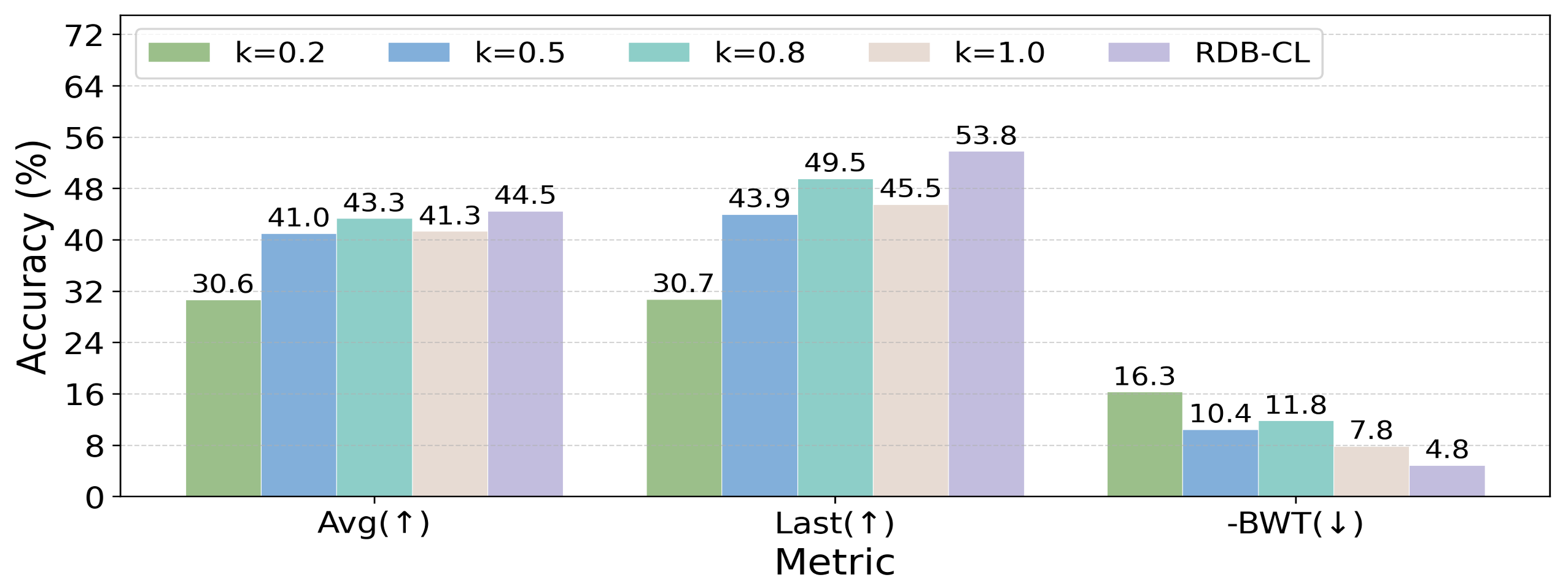}

    \captionof{figure}{Performance comparison of static KL with fixed $\beta$ (scaled by $k$ from $\beta_0 = 0.15$) and dynamic KL with $\beta_{LRC}$.}
    \label{fig:k_bars}
  \end{minipage}
\end{figure}

\paragraph{Reasoning Confidence vs. Answer Confidence.}

To validate the advantage of reasoning-level signals, we compare \emph{answer-level} and \emph{reasoning-level} confidence under identical experimental settings. In both variants, the KL constraint is modulated by calculated answer confidence and reasoning confidence respectively during training. As shown in \cref{tab:answer_result}, the reasoning-level delivers consistent gains across all reported metrics(\textit{BwT}: -4.98), whereas the answer-level exacerbates forgetting(\textit{BwT}: -8.68) relative to the GRPO baseline.
These results align with the observation in \cref{sebsec:confidence_study}: answer confidence becomes noisy on new-task (OOD) samples, triggering effectively random intervention. While this may transiently increase plasticity, such mistimed relaxation near the old-task manifolds induces larger drifts, resulting in severe catastrophic forgetting. In contrast, reasoning confidence provides reliable guidance.

\paragraph{Static KL vs. Dynamic KL.}

To clarify whether our method essentially amounts to re-tuning the hyperparameter $\beta$ to control the KL regularizer statically, we compare against a family of baselines with static $\beta$. Concretely, starting from original $\beta_0 = 0.15$, we scale the coefficient by factors $k$ during training, $\beta_{static} = k\beta_0$, and report results in \cref{fig:k_bars}. As shown, our \emph{reasoning-portability-aware} modulation via $\beta_{LRC}$ consistently outperforms all fixed settings across metrics. 
The curve for $k=0.2$ is omitted to preserve figure readability, as this setting typically causes unstable dynamics characterized by extreme shifts and stagnated updates.
Overall, our method yields the most stable yet adaptive training dynamics, indicating stronger robustness against forgetting than any static choice.
To further intuitively illustrate this stability, we visualize the KL divergence training dynamics in Appendix \ref{app:train_details}, \cref{fig:kl_dynamics}, confirming that RDB-CL achieves the most robust convergence without extreme feature drift.

\subsection{Diversity Retention for RLVR in CL}

\begin{wrapfigure}[13]{r}{0.47\linewidth}
\vspace{-12pt}
  \centering
  \includegraphics[width=\linewidth]{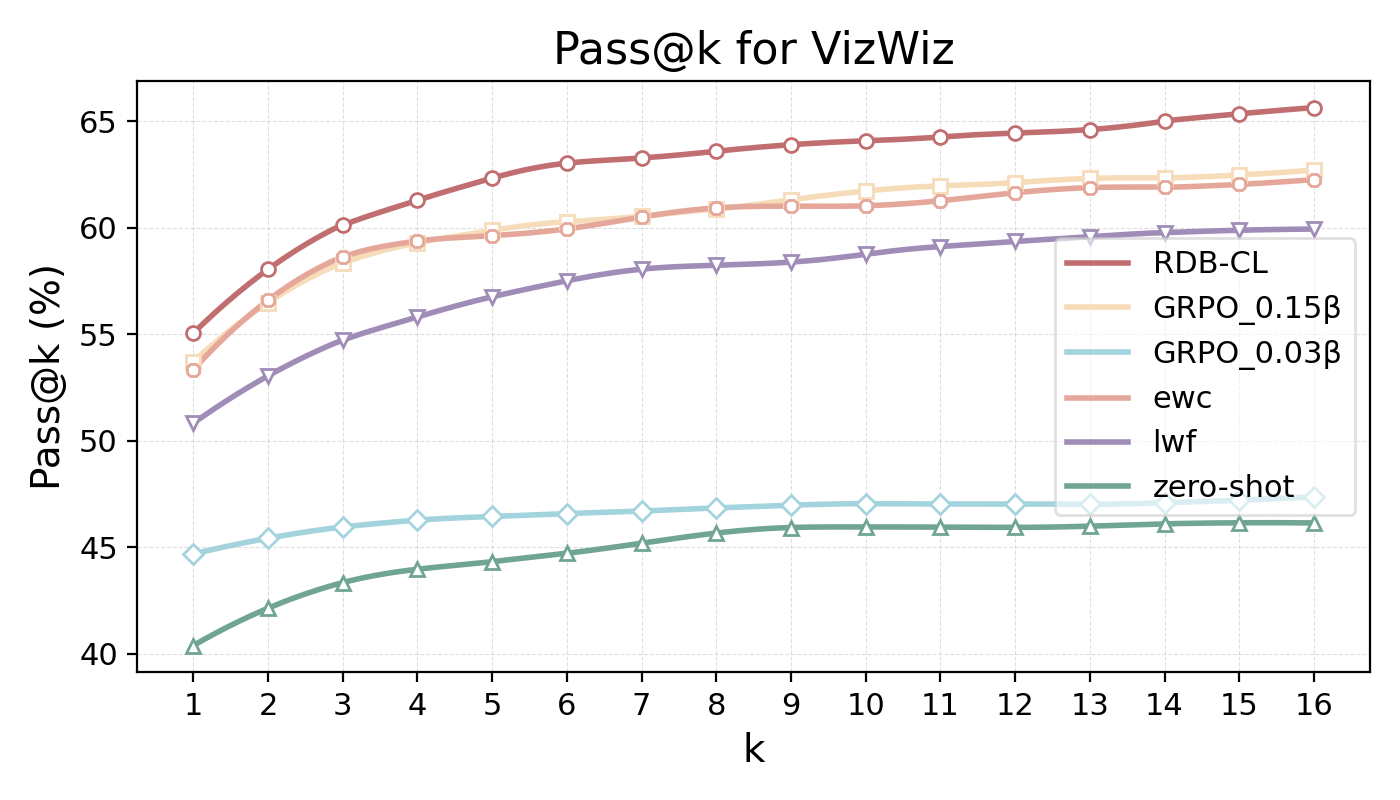}
  \vspace{-10pt}
  \caption{
  Comparison of pass@k results on the first task VizWiz.
  }
  \label{fig:passk}
\end{wrapfigure}

Unlike SFT, RL inherently preserves prior knowledge~\citep{RLRazor, retainingdoingroleonpolicy}, yet it is prone to plasticity loss from exploration collapse~\citep{negativereinforcement, entropy}, a problem further amplified in CL. We observed that under a weak or absent KL constraint, plasticity diminishes even to zero: gradient magnitudes become insufficient to move the policy away from its initial mode, and exploration stalls.
From this perspective, the loss of diversity in the initial policy distribution constitutes a latent form of catastrophic forgetting: since both continual exploration and the initialization for downstream tasks upper-bound RL's attainable performance~\citep{wise-ft,really}, preserving diversity is critical in the CL regime.

We use \textit{Pass@k} as an empirical proxy for generative diversity~\citep{really}, and deliberately evaluate it on the first training node as a controlled probe (\cref{fig:passk}). Higher \textit{Pass@k} for increasing $k$ indicates that the model produces a wider set of distinct hypotheses rather than repeatedly sampling from a small number of modes, which induces more exploration. 
GRPO’s \textit{Pass@k} curve grows only marginally with $k$, indicating diversity exhaustion that restricts exploration and thus plasticity across sequential tasks.
In contrast, RDB-CL maintains a \textit{consistently rising} \textit{Pass@k}, suggesting that reasoning-portability-aware updates allocate exploration to non-transferable cases and preserve diversity, thereby enabling stronger potential for continual adaptation.

\subsection{Hyperparameter Discussion}
\paragraph{Reasoning Score Threshold $\tau$.}
The threshold $\tau$ serves as a key component in RDB-CL, providing a qualitative categorization of reasoning portability. Samples with reasoning confidence below $\tau$ are treated as NRP and encouraged to explore, while those above 
$\tau$ are considered PRP and guided with stronger regularization. We conduct a hyperparameter study across different $\tau$ values, with results summarized in~\cref{tab:threshold_result}. Across a wide and reasonable range, RDB-CL exhibits consistently strong performance and stability improvements, demonstrating substantial robustness to the choice of $\tau$. Even small thresholds, where only a small number of samples are allowed to explore, still bring noticeable gains, suggesting the importance of these exploratory cases in shaping new subspaces. Extremely large $\tau$ may misclassify PRP samples as NRP, leading to unnecessary exploration and increased forgetting, as the distribution shown in \cref{fig:conf_dist}. Nevertheless, the method maintains reliable performance overall. These results collectively show that RDB-CL remains effective across diverse $\tau$ configurations, reinforcing the robustness of our reasoning-guided design.

\paragraph{Minimum KL constraint ${clip}_{min}$.}
We analyze the effect of varying $clip_{min}$, which enforces a lower bound on the KL penalty and is intended to improve training stability. As the result shown in \cref{tab:clip_result}, even when no minimum KL constraint is applied ($clip_{min}=0$), RDB-CL still yields notable gains over GRPO and mitigates forgetting. This indicates that the primary mechanism of forgetting reduction does not stem from the minimum KL constraint itself but reasoning-guided modulation. Nevertheless, a nonzero $clip_{min}$ does contribute to overall stability. Extremely large values, however, overly restrict exploration for low-confidence samples, leading to under-optimization as \textit{Finetune} for $clip_{min}=0.3$ is inferior.
Overall, RDB-CL demonstrates strong robustness to the choice of $clip_{min}$, with a wide range of values consistently outperforming the baseline.
 
\begin{table}[t]
  \centering
  \begin{minipage}[b]{0.45\linewidth}
    \centering
    \caption{Performance comparison across different $\tau$, with $clip_{min}$ fixed at $0.2$ and $\tau=0.7$ used in our main experiments.}
    \label{tab:threshold_result} 
    \renewcommand{\arraystretch}{1.1} 
\centering
\small
\begin{tabular}
 {p{1.1cm}|cccc}
\toprule
Method   & \textit{Avg.}  & \textit{Last} & \textit{Finetune} & \textit{BwT}\\ 
\midrule
GRPO          & 41.32  & 45.40  & 50.63 & -7.84\\
$\tau=0.5$     & 44.06  & 54.11  & \textbf{57.07} & -4.43\\
$\tau=0.6$  & \textbf{45.34}  & \textbf{54.35}  & 56.22 & \textbf{-2.80}\\
$\tau=0.7$  & 44.49  & 53.79  & 57.01 & -4.83\\
$\tau=0.8$  & 43.32  & 48.21  & 52.23 & -6.56\\
\bottomrule
\end{tabular}

  \end{minipage}\hfill
  \begin{minipage}[b]{0.51\linewidth}
    \centering
    \caption{Performance comparison across different $clip_{min}$, with $\tau$ fixed at $0.7$ and $clip_{min}=0.2$ used in main experiments.}
    \label{tab:clip_result} 
    \renewcommand{\arraystretch}{1.1} 
\centering
\small
\begin{tabular}{l|cccc}
\toprule
Method   & \textit{Avg.}  & \textit{Last} & \textit{Finetune} & \textit{BwT}\\ 
\midrule
GRPO          & 41.32  & 45.40  & 50.63 & -7.84\\
$clip_{min}=0.0$     & 43.21  & 51.27  & 55.57 & -5.67\\
$clip_{min}=0.1$  & \textbf{45.09}  & \textbf{54.50}  & \textbf{58.20} & -5.14\\
$clip_{min}=0.2$  & 44.49  & 53.79  & 57.01 & -4.83\\
$clip_{min}=0.3$  & 43.67  & 52.83  & 53.37 & \textbf{-4.66}\\
\bottomrule
\end{tabular} 
  \end{minipage}
\end{table}


\label{subsec:analysis_scale}

\section{Conclusion}
\label{sec:conclusion}

Motivated by the growing reasoning capabilities of MLLMs and the mismatch between traditional CL constraints and the new paradigm, we seek a reliable signal to guide continual adaptation. We empirically show that reasoning-level confidence tracks cross-task 
transferability more reliably than answer-level under distribution shift, and use it as a per-sample proxy for \emph{reasoning portability}. Building on this, we propose Reasoning-based Dynamic Balance Continual Learning (RDB-CL), which integrates naturally 
with the KL regularizer in RLVR by turning it into a 
portability-gated constraint, achieving a plasticity--stability trade-off. Experiments on VQA benchmarks show RDB-CL outperforms GRPO and classical CL methods, remains robust across task orders 
and larger scales, and is further validated by ablations, diversity measurements, and hyperparameter analyses.

\paragraph{Limitations and Future Work.} 
While we show the effectiveness of RDB-CL, several limitations remain. (1) Our evaluation is constrained by computational budget; future work can extend to a wider range of tasks, domains, and text-only LLM continual learning settings.
(2) Reasoning confidence is only one possible portability estimator, and efficient alternatives such as calibrated uncertainty, verifier-based signals, self-consistency remain to be explored.
(3) The current sample-level scalar modulation is simple but heuristic, and finer-grained schemes or more comprehensive modulation with learned threshold offer a promising direction for more principled control of reasoning drift.

\bibliographystyle{plainnat}
\bibliography{main}

@String(NeurIPS = {Adv. Neural Inform. Process. Syst.})

@String(NeurIPS = {NeurIPS})

@article{VLM-CLSurvey,
  title={Continual Learning for VLMs: A Survey and Taxonomy Beyond Forgetting},
  author={Liu, Yuyang and Hong, Qiuhe and Huang, Linlan and Gomez-Villa, Alexandra and Goswami, Dipam and Liu, Xialei and van de Weijer, Joost and Tian, Yonghong},
  journal={arXiv preprint arXiv:2508.04227},
  year={2025}
}

@inproceedings{vqacl,
  title={Vqacl: A novel visual question answering continual learning setting},
  author={Zhang, Xi and Zhang, Feifei and Xu, Changsheng},
  booktitle={Proceedings of the IEEE/CVF Conference on Computer Vision and Pattern Recognition},
  pages={19102--19112},
  year={2023}
}

@article{mspt,
  title={Continual multimodal knowledge graph construction},
  author={Chen, Xiang and Zhang, Jintian and Wang, Xiaohan and Zhang, Ningyu and Wu, Tongtong and Wang, Yuxiang and Wang, Yongheng and Chen, Huajun},
  journal={arXiv preprint arXiv:2305.08698},
  year={2023}
}

@inproceedings{incclip,
  title={Generative negative text replay for continual vision-language pretraining},
  author={Yan, Shipeng and Hong, Lanqing and Xu, Hang and Han, Jianhua and Tuytelaars, Tinne and Li, Zhenguo and He, Xuming},
  booktitle={European Conference on Computer Vision},
  pages={22--38},
  year={2022},
  organization={Springer}
}

@inproceedings{dualteacher,
  title={Select and distill: Selective dual-teacher knowledge transfer for continual learning on vision-language models},
  author={Yu, Yu-Chu and Huang, Chi-Pin and Chen, Jr-Jen and Chang, Kai-Po and Lai, Yung-Hsuan and Yang, Fu-En and Wang, Yu-Chiang Frank},
  booktitle={European Conference on Computer Vision},
  pages={219--236},
  year={2024},
  organization={Springer}
}

@inproceedings{scd,
  title={Multi-domain lifelong visual question answering via self-critical distillation},
  author={Lao, Mingrui and Pu, Nan and Liu, Yu and Zhong, Zhun and Bakker, Erwin M and Sebe, Nicu and Lew, Michael S},
  booktitle={Proceedings of the 31st ACM International Conference on Multimedia},
  pages={4747--4758},
  year={2023}
}

@article{sgcl,
  title={Exploiting the semantic knowledge of pre-trained text-encoders for continual learning},
  author={Yu, Lu and Tao, Zhe and Goswami, Dipam and Yao, Hantao and Twardowski, Bart{\l}omiej and Van de Weijer, Joost and Xu, Changsheng},
  journal={arXiv preprint arXiv:2408.01076},
  year={2024}
}

@inproceedings{zscl,
  title={Preventing zero-shot transfer degradation in continual learning of vision-language models},
  author={Zheng, Zangwei and Ma, Mingyuan and Wang, Kai and Qin, Ziheng and Yue, Xiangyu and You, Yang},
  booktitle={Proceedings of the IEEE/CVF international conference on computer vision},
  pages={19125--19136},
  year={2023}
}

@inproceedings{coin,
 author = {Chen, Cheng and Zhu, Junchen and Luo, Xu and Shen, Heng Tao and Song, Jingkuan and Gao, Lianli},
 booktitle = {Advances in Neural Information Processing Systems},
 editor = {A. Globerson and L. Mackey and D. Belgrave and A. Fan and U. Paquet and J. Tomczak and C. Zhang},
 pages = {57817--57840},
 publisher = {Curran Associates, Inc.},
 title = {CoIN: A Benchmark of Continual Instruction Tuning for Multimodel Large Language Models},
 url = {https://proceedings.neurips.cc/paper_files/paper/2024/file/6a45500d9eda640deed90d8a62742be5-Paper-Datasets_and_Benchmarks_Track.pdf},
 volume = {37},
 year = {2024}
}

@ARTICLE{proof,
  author={Zhou, Da-Wei and Zhang, Yuanhan and Wang, Yan and Ning, Jingyi and Ye, Han-Jia and Zhan, De-Chuan and Liu, Ziwei},
  journal={IEEE Transactions on Pattern Analysis and Machine Intelligence}, 
  title={Learning Without Forgetting for Vision-Language Models}, 
  year={2025},
  volume={47},
  number={6},
  pages={4489-4504},
  doi={10.1109/TPAMI.2025.3540889}}

@unknown{rail,
author = {Xu, Yicheng and Chen, Yuxin and Nie, Jiahao and Wang, Yusong and Zhuang, Huiping and Okumura, Manabu},
year = {2024},
month = {06},
pages = {},
title = {Advancing Cross-domain Discriminability in Continual Learning of Vison-Language Models},
doi = {10.48550/arXiv.2406.18868}
}

@article{hide,
  title={Hide-llava: Hierarchical decoupling for continual instruction tuning of multimodal large language model},
  author={Guo, Haiyang and Zeng, Fanhu and Xiang, Ziwei and Zhu, Fei and Wang, Da-Han and Zhang, Xu-Yao and Liu, Cheng-Lin},
  journal={arXiv preprint arXiv:2503.12941},
  year={2025}
}

@article{deepseek,
  title={Deepseek-v3 technical report},
  author={Liu, Aixin and Feng, Bei and Xue, Bing and Wang, Bingxuan and Wu, Bochao and Lu, Chengda and Zhao, Chenggang and Deng, Chengqi and Zhang, Chenyu and Ruan, Chong and others},
  journal={arXiv preprint arXiv:2412.19437},
  year={2024}
}

@misc{deepseekmath,
      title={DeepSeekMath: Pushing the Limits of Mathematical Reasoning in Open Language Models}, 
      author={Zhihong Shao and Peiyi Wang and Qihao Zhu and Runxin Xu and Junxiao Song and Xiao Bi and Haowei Zhang and Mingchuan Zhang and Y. K. Li and Y. Wu and Daya Guo},
      year={2024},
      eprint={2402.03300},
      archivePrefix={arXiv},
      primaryClass={cs.CL},
      url={https://arxiv.org/abs/2402.03300}, 
}

@misc{RLRazor,
      title={RL's Razor: Why Online Reinforcement Learning Forgets Less}, 
      author={Idan Shenfeld and Jyothish Pari and Pulkit Agrawal},
      year={2025},
      eprint={2509.04259},
      archivePrefix={arXiv},
      primaryClass={cs.LG},
      url={https://arxiv.org/abs/2509.04259}, 
}

@misc{RL-MLLM-forgetting,
      title={Why Reinforcement Fine-Tuning Enables MLLMs Preserve Prior Knowledge Better: A Data Perspective}, 
      author={Zhihao Zhang and Qiaole Dong and Qi Zhang and Jun Zhao and Enyu Zhou and Zhiheng Xi and Senjie Jin and Xiaoran Fan and Yuhao Zhou and Mingqi Wu and Yanwei Fu and Tao Ji and Tao Gui and Xuanjing Huang and Kai Chen},
      year={2025},
      eprint={2506.23508},
      archivePrefix={arXiv},
      primaryClass={cs.CL},
      url={https://arxiv.org/abs/2506.23508}, 
}

@misc{wise-ft,
      title={Weight Ensembling Improves Reasoning in Language Models}, 
      author={Xingyu Dang and Christina Baek and Kaiyue Wen and Zico Kolter and Aditi Raghunathan},
      year={2025},
      eprint={2504.10478},
      archivePrefix={arXiv},
      primaryClass={cs.LG},
      url={https://arxiv.org/abs/2504.10478}, 
}

@misc{ptrue,
      title={Language Models (Mostly) Know What They Know}, 
      author={Saurav Kadavath and Tom Conerly and Amanda Askell and Tom Henighan and Dawn Drain and Ethan Perez and Nicholas Schiefer and Zac Hatfield-Dodds and Nova DasSarma and Eli Tran-Johnson and Scott Johnston and Sheer El-Showk and Andy Jones and Nelson Elhage and Tristan Hume and Anna Chen and Yuntao Bai and Sam Bowman and Stanislav Fort and Deep Ganguli and Danny Hernandez and Josh Jacobson and Jackson Kernion and Shauna Kravec and Liane Lovitt and Kamal Ndousse and Catherine Olsson and Sam Ringer and Dario Amodei and Tom Brown and Jack Clark and Nicholas Joseph and Ben Mann and Sam McCandlish and Chris Olah and Jared Kaplan},
      year={2022},
      eprint={2207.05221},
      archivePrefix={arXiv},
      primaryClass={cs.CL},
      url={https://arxiv.org/abs/2207.05221}, 
}

@inproceedings{rlhf,
   title={Confidence Improves Self-Consistency in LLMs},
   url={http://dx.doi.org/10.18653/v1/2025.findings-acl.1030},
   DOI={10.18653/v1/2025.findings-acl.1030},
   booktitle={Findings of the Association for Computational Linguistics: ACL 2025},
   publisher={Association for Computational Linguistics},
   author={Taubenfeld, Amir and Sheffer, Tom and Ofek, Eran and Feder, Amir and Goldstein, Ariel and Gekhman, Zorik and Yona, Gal},
   year={2025},
   pages={20090–20111} }

@inproceedings{iconqa,
    title = {IconQA: A New Benchmark for Abstract Diagram Understanding and Visual Language Reasoning},
    author = {Lu, Pan and Qiu, Liang and Chen, Jiaqi and Xia, Tony and Zhao, Yizhou and Zhang, Wei and Yu, Zhou and Liang, Xiaodan and Zhu, Song-Chun},
    booktitle = {The 35th Conference on Neural Information Processing Systems (NeurIPS 2021) Track on Datasets and Benchmarks},
    year = {2021}
}

@misc{vizwiz,
      title={VizWiz Grand Challenge: Answering Visual Questions from Blind People}, 
      author={Danna Gurari and Qing Li and Abigale J. Stangl and Anhong Guo and Chi Lin and Kristen Grauman and Jiebo Luo and Jeffrey P. Bigham},
      year={2018},
      eprint={1802.08218},
      archivePrefix={arXiv},
      primaryClass={cs.CV},
      url={https://arxiv.org/abs/1802.08218}, 
}

@INPROCEEDINGS{imagenet,
  author={Deng, Jia and Dong, Wei and Socher, Richard and Li, Li-Jia and Kai Li and Li Fei-Fei},
  booktitle={2009 IEEE Conference on Computer Vision and Pattern Recognition}, 
  title={ImageNet: A large-scale hierarchical image database}, 
  year={2009},
  volume={},
  number={},
  pages={248-255},
  doi={10.1109/CVPR.2009.5206848}}

@article{ewc,
   title={Overcoming catastrophic forgetting in neural networks},
   volume={114},
   ISSN={1091-6490},
   url={http://dx.doi.org/10.1073/pnas.1611835114},
   DOI={10.1073/pnas.1611835114},
   number={13},
   journal={Proceedings of the National Academy of Sciences},
   publisher={Proceedings of the National Academy of Sciences},
   author={Kirkpatrick, James and Pascanu, Razvan and Rabinowitz, Neil and Veness, Joel and Desjardins, Guillaume and Rusu, Andrei A. and Milan, Kieran and Quan, John and Ramalho, Tiago and Grabska-Barwinska, Agnieszka and Hassabis, Demis and Clopath, Claudia and Kumaran, Dharshan and Hadsell, Raia},
   year={2017},
   month=mar, pages={3521–3526} }

@misc{lwf,
      title={Learning without Forgetting}, 
      author={Zhizhong Li and Derek Hoiem},
      year={2017},
      eprint={1606.09282},
      archivePrefix={arXiv},
      primaryClass={cs.CV},
      url={https://arxiv.org/abs/1606.09282}, 
}

@inproceedings{bwt,
  title={Gradient Episodic Memory for Continual Learning},
  author={Lopez-Paz, David and Ranzato, Marc'Aurelio},
  booktitle={NeurIPS},
  year={2017},
  url={https://arxiv.org/abs/1706.08840},
  doi={10.48550/arXiv.1706.08840}
}

@misc{qwen2.5-VL,
    title = {Qwen2.5-VL},
    url = {https://qwenlm.github.io/blog/qwen2.5-vl/},
    author = {Qwen Team},
    month = {January},
    year = {2025}
}

@article{tan2025reason,
  title={Reason-RFT: Reinforcement Fine-Tuning for Visual Reasoning},
  author={Tan, Huajie and Ji, Yuheng and Hao, Xiaoshuai and Lin, Minglan and Wang, Pengwei and Wang, Zhongyuan and Zhang, Shanghang},
  journal={arXiv preprint arXiv:2503.20752},
  year={2025}
}

@inproceedings{vllm,
  title={Efficient Memory Management for Large Language Model Serving with PagedAttention},
  author={Woosuk Kwon and Zhuohan Li and Siyuan Zhuang and Ying Sheng and Lianmin Zheng and Cody Hao Yu and Joseph E. Gonzalez and Hao Zhang and Ion Stoica},
  booktitle={Proceedings of the ACM SIGOPS 29th Symposium on Operating Systems Principles},
  year={2023}
}

@misc{retainingdoingroleonpolicy,
      title={Retaining by Doing: The Role of On-Policy Data in Mitigating Forgetting}, 
      author={Howard Chen and Noam Razin and Karthik Narasimhan and Danqi Chen},
      year={2025},
      eprint={2510.18874},
      archivePrefix={arXiv},
      primaryClass={cs.LG},
      url={https://arxiv.org/abs/2510.18874}, 
}

@misc{negativereinforcement,
      title={The Surprising Effectiveness of Negative Reinforcement in LLM Reasoning}, 
      author={Xinyu Zhu and Mengzhou Xia and Zhepei Wei and Wei-Lin Chen and Danqi Chen and Yu Meng},
      year={2025},
      eprint={2506.01347},
      archivePrefix={arXiv},
      primaryClass={cs.CL},
      url={https://arxiv.org/abs/2506.01347}, 
}

@misc{really,
      title={Does Reinforcement Learning Really Incentivize Reasoning Capacity in LLMs Beyond the Base Model?}, 
      author={Yang Yue and Zhiqi Chen and Rui Lu and Andrew Zhao and Zhaokai Wang and Yang Yue and Shiji Song and Gao Huang},
      year={2025},
      eprint={2504.13837},
      archivePrefix={arXiv},
      primaryClass={cs.AI},
      url={https://arxiv.org/abs/2504.13837}, 
}

@misc{entropy,
      title={The Entropy Mechanism of Reinforcement Learning for Reasoning Language Models}, 
      author={Ganqu Cui and Yuchen Zhang and Jiacheng Chen and Lifan Yuan and Zhi Wang and Yuxin Zuo and Haozhan Li and Yuchen Fan and Huayu Chen and Weize Chen and Zhiyuan Liu and Hao Peng and Lei Bai and Wanli Ouyang and Yu Cheng and Bowen Zhou and Ning Ding},
      year={2025},
      eprint={2505.22617},
      archivePrefix={arXiv},
      primaryClass={cs.LG},
      url={https://arxiv.org/abs/2505.22617}, 
}

@misc{tulu3,
      title={Tulu 3: Pushing Frontiers in Open Language Model Post-Training}, 
      author={Nathan Lambert and Jacob Morrison and Valentina Pyatkin and Shengyi Huang and Hamish Ivison and Faeze Brahman and Lester James V. Miranda and Alisa Liu and Nouha Dziri and Shane Lyu and Yuling Gu and Saumya Malik and Victoria Graf and Jena D. Hwang and Jiangjiang Yang and Ronan Le Bras and Oyvind Tafjord and Chris Wilhelm and Luca Soldaini and Noah A. Smith and Yizhong Wang and Pradeep Dasigi and Hannaneh Hajishirzi},
      year={2025},
      eprint={2411.15124},
      archivePrefix={arXiv},
      primaryClass={cs.CL},
      url={https://arxiv.org/abs/2411.15124}, 
}

@misc{dapo,
      title={DAPO: An Open-Source LLM Reinforcement Learning System at Scale}, 
      author={Qiying Yu and Zheng Zhang and Ruofei Zhu and Yufeng Yuan and Xiaochen Zuo and Yu Yue and Weinan Dai and Tiantian Fan and Gaohong Liu and Lingjun Liu and Xin Liu and Haibin Lin and Zhiqi Lin and Bole Ma and Guangming Sheng and Yuxuan Tong and Chi Zhang and Mofan Zhang and Wang Zhang and Hang Zhu and Jinhua Zhu and Jiaze Chen and Jiangjie Chen and Chengyi Wang and Hongli Yu and Yuxuan Song and Xiangpeng Wei and Hao Zhou and Jingjing Liu and Wei-Ying Ma and Ya-Qin Zhang and Lin Yan and Mu Qiao and Yonghui Wu and Mingxuan Wang},
      year={2025},
      eprint={2503.14476},
      archivePrefix={arXiv},
      primaryClass={cs.LG},
      url={https://arxiv.org/abs/2503.14476}, 
}

@misc{r1_vl,
      title={R1-VL: Learning to Reason with Multimodal Large Language Models via Step-wise Group Relative Policy Optimization}, 
      author={Jingyi Zhang and Jiaxing Huang and Huanjin Yao and Shunyu Liu and Xikun Zhang and Shijian Lu and Dacheng Tao},
      year={2025},
      eprint={2503.12937},
      archivePrefix={arXiv},
      primaryClass={cs.AI},
      url={https://arxiv.org/abs/2503.12937}, 
}

@misc{mmeureka,
      title={MM-Eureka: Exploring the Frontiers of Multimodal Reasoning with Rule-based Reinforcement Learning}, 
      author={Fanqing Meng and Lingxiao Du and Zongkai Liu and Zhixiang Zhou and Quanfeng Lu and Daocheng Fu and Tiancheng Han and Botian Shi and Wenhai Wang and Junjun He and Kaipeng Zhang and Ping Luo and Yu Qiao and Qiaosheng Zhang and Wenqi Shao},
      year={2025},
      eprint={2503.07365},
      archivePrefix={arXiv},
      primaryClass={cs.CV},
      url={https://arxiv.org/abs/2503.07365}, 
}

@misc{deepeyes,
      title={DeepEyes: Incentivizing "Thinking with Images" via Reinforcement Learning}, 
      author={Ziwei Zheng and Michael Yang and Jack Hong and Chenxiao Zhao and Guohai Xu and Le Yang and Chao Shen and Xing Yu},
      year={2025},
      eprint={2505.14362},
      archivePrefix={arXiv},
      primaryClass={cs.CV},
      url={https://arxiv.org/abs/2505.14362}, 
}

@misc{mvot,
      title={Imagine while Reasoning in Space: Multimodal Visualization-of-Thought}, 
      author={Chengzu Li and Wenshan Wu and Huanyu Zhang and Yan Xia and Shaoguang Mao and Li Dong and Ivan Vulić and Furu Wei},
      year={2025},
      eprint={2501.07542},
      archivePrefix={arXiv},
      primaryClass={cs.CL},
      url={https://arxiv.org/abs/2501.07542}, 
}

@misc{diki,
      title={Mind the Interference: Retaining Pre-trained Knowledge in Parameter Efficient Continual Learning of Vision-Language Models}, 
      author={Longxiang Tang and Zhuotao Tian and Kai Li and Chunming He and Hantao Zhou and Hengshuang Zhao and Xiu Li and Jiaya Jia},
      year={2024},
      eprint={2407.05342},
      archivePrefix={arXiv},
      primaryClass={cs.CV},
      url={https://arxiv.org/abs/2407.05342}, 
}

@misc{moeadapters,
      title={Boosting Continual Learning of Vision-Language Models via Mixture-of-Experts Adapters}, 
      author={Jiazuo Yu and Yunzhi Zhuge and Lu Zhang and Ping Hu and Dong Wang and Huchuan Lu and You He},
      year={2024},
      eprint={2403.11549},
      archivePrefix={arXiv},
      primaryClass={cs.CV},
      url={https://arxiv.org/abs/2403.11549}, 
}

@misc{adaptinfty,
      title={Adapt-$\infty$: Scalable Continual Multimodal Instruction Tuning via Dynamic Data Selection}, 
      author={Adyasha Maharana and Jaehong Yoon and Tianlong Chen and Mohit Bansal},
      year={2025},
      eprint={2410.10636},
      archivePrefix={arXiv},
      primaryClass={cs.LG},
      url={https://arxiv.org/abs/2410.10636}, 
}

@misc{mllmcl,
      title={MLLM-CL: Continual Learning for Multimodal Large Language Models}, 
      author={Hongbo Zhao and Fei Zhu and Haiyang Guo and Meng Wang and Rundong Wang and Gaofeng Meng and Zhaoxiang Zhang},
      year={2025},
      eprint={2506.05453},
      archivePrefix={arXiv},
      primaryClass={cs.CL},
      url={https://arxiv.org/abs/2506.05453}, 
}

@inproceedings{
cppo,
title={{CPPO}: Continual Learning for Reinforcement Learning with Human Feedback},
author={Han Zhang and Yu Lei and Lin Gui and Min Yang and Yulan He and Hui Wang and Ruifeng Xu},
booktitle={The Twelfth International Conference on Learning Representations},
year={2024},
url={https://openreview.net/forum?id=86zAUE80pP}
}

@misc{mtcore,
      title={Multi-granularity Knowledge Transfer for Continual Reinforcement Learning}, 
      author={Chaofan Pan and Lingfei Ren and Yihui Feng and Linbo Xiong and Wei Wei and Yonghao Li and Xin Yang},
      year={2025},
      eprint={2401.15098},
      archivePrefix={arXiv},
      primaryClass={cs.LG},
      url={https://arxiv.org/abs/2401.15098}, 
}

@misc{internalconsistencyselffeedbacklargesurvey,
      title={Internal Consistency and Self-Feedback in Large Language Models: A Survey}, 
      author={Xun Liang and Shichao Song and Zifan Zheng and Hanyu Wang and Qingchen Yu and Xunkai Li and Rong-Hua Li and Yi Wang and Zhonghao Wang and Feiyu Xiong and Zhiyu Li},
      year={2024},
      eprint={2407.14507},
      archivePrefix={arXiv},
      primaryClass={cs.CL},
      url={https://arxiv.org/abs/2407.14507}, 
}

@misc{selfconsistency,
      title={Self-Consistency Improves Chain of Thought Reasoning in Language Models}, 
      author={Xuezhi Wang and Jason Wei and Dale Schuurmans and Quoc Le and Ed Chi and Sharan Narang and Aakanksha Chowdhery and Denny Zhou},
      year={2023},
      eprint={2203.11171},
      archivePrefix={arXiv},
      primaryClass={cs.CL},
      url={https://arxiv.org/abs/2203.11171}, 
}

@misc{trustscore,
      title={TrustScore: Reference-Free Evaluation of LLM Response Trustworthiness}, 
      author={Danna Zheng and Danyang Liu and Mirella Lapata and Jeff Z. Pan},
      year={2024},
      eprint={2402.12545},
      archivePrefix={arXiv},
      primaryClass={cs.CL},
      url={https://arxiv.org/abs/2402.12545}, 
}

@misc{datauncertainty,
      title={Uncertainty Aware Learning for Language Model Alignment}, 
      author={Yikun Wang and Rui Zheng and Liang Ding and Qi Zhang and Dahua Lin and Dacheng Tao},
      year={2024},
      eprint={2406.04854},
      archivePrefix={arXiv},
      primaryClass={cs.CL},
      url={https://arxiv.org/abs/2406.04854}, 
}

@misc{uncertaintyVLM,
      title={Uncertainty-Aware Evaluation for Vision-Language Models}, 
      author={Vasily Kostumov and Bulat Nutfullin and Oleg Pilipenko and Eugene Ilyushin},
      year={2024},
      eprint={2402.14418},
      archivePrefix={arXiv},
      primaryClass={cs.CV},
      url={https://arxiv.org/abs/2402.14418}, 
}

@misc{huang2022largelanguagemodelsselfimprove,
      title={Large Language Models Can Self-Improve}, 
      author={Jiaxin Huang and Shixiang Shane Gu and Le Hou and Yuexin Wu and Xuezhi Wang and Hongkun Yu and Jiawei Han},
      year={2022},
      eprint={2210.11610},
      archivePrefix={arXiv},
      primaryClass={cs.CL},
      url={https://arxiv.org/abs/2210.11610}, 
}

@misc{prasad2025selfconsistencypreferenceoptimization,
      title={Self-Consistency Preference Optimization}, 
      author={Archiki Prasad and Weizhe Yuan and Richard Yuanzhe Pang and Jing Xu and Maryam Fazel-Zarandi and Mohit Bansal and Sainbayar Sukhbaatar and Jason Weston and Jane Yu},
      year={2025},
      eprint={2411.04109},
      archivePrefix={arXiv},
      primaryClass={cs.CL},
      url={https://arxiv.org/abs/2411.04109}, 
}

@misc{reasoningconfidence,
      title={Self-Training Large Language Models with Confident Reasoning}, 
      author={Hyosoon Jang and Yunhui Jang and Sungjae Lee and Jungseul Ok and Sungsoo Ahn},
      year={2025},
      eprint={2505.17454},
      archivePrefix={arXiv},
      primaryClass={cs.LG},
      url={https://arxiv.org/abs/2505.17454}, 
}

@misc{ucas,
      title={Unlocking Exploration in RLVR: Uncertainty-aware Advantage Shaping for Deeper Reasoning}, 
      author={Can Xie and Ruotong Pan and Xiangyu Wu and Yunfei Zhang and Jiayi Fu and Tingting Gao and Guorui Zhou},
      year={2025},
      eprint={2510.10649},
      archivePrefix={arXiv},
      primaryClass={cs.AI},
      url={https://arxiv.org/abs/2510.10649}, 
}

@misc{llm,
      title={Language Models are Few-Shot Learners}, 
      author={Tom B. Brown and Benjamin Mann and Nick Ryder and Melanie Subbiah and Jared Kaplan and Prafulla Dhariwal and Arvind Neelakantan and Pranav Shyam and Girish Sastry and Amanda Askell and Sandhini Agarwal and Ariel Herbert-Voss and Gretchen Krueger and Tom Henighan and Rewon Child and Aditya Ramesh and Daniel M. Ziegler and Jeffrey Wu and Clemens Winter and Christopher Hesse and Mark Chen and Eric Sigler and Mateusz Litwin and Scott Gray and Benjamin Chess and Jack Clark and Christopher Berner and Sam McCandlish and Alec Radford and Ilya Sutskever and Dario Amodei},
      year={2020},
      eprint={2005.14165},
      archivePrefix={arXiv},
      primaryClass={cs.CL},
      url={https://arxiv.org/abs/2005.14165}, 
}

@article{clsurvey,
   title={A continual learning survey: Defying forgetting in classification tasks},
   ISSN={1939-3539},
   url={http://dx.doi.org/10.1109/TPAMI.2021.3057446},
   DOI={10.1109/tpami.2021.3057446},
   journal={IEEE Transactions on Pattern Analysis and Machine Intelligence},
   publisher={Institute of Electrical and Electronics Engineers (IEEE)},
   author={Delange, Matthias and Aljundi, Rahaf and Masana, Marc and Parisot, Sarah and Jia, Xu and Leonardis, Ales and Slabaugh, Greg and Tuytelaars, Tinne},
   year={2021},
   pages={1–1} }

@article{clsurvey2,
title = {Continual lifelong learning with neural networks: A review},
journal = {Neural Networks},
volume = {113},
pages = {54-71},
year = {2019},
issn = {0893-6080},
doi = {https://doi.org/10.1016/j.neunet.2019.01.012},
url = {https://www.sciencedirect.com/science/article/pii/S0893608019300231},
author = {German I. Parisi and Ronald Kemker and Jose L. Part and Christopher Kanan and Stefan Wermter}
}

@article{kl,
  title={On information and sufficiency},
  author={Kullback, Solomon and Leibler, Richard A},
  journal={The annals of mathematical statistics},
  volume={22},
  number={1},
  pages={79--86},
  year={1951},
  publisher={JSTOR}
}

@misc{instructblip,
      title={InstructBLIP: Towards General-purpose Vision-Language Models with Instruction Tuning}, 
      author={Wenliang Dai and Junnan Li and Dongxu Li and Anthony Meng Huat Tiong and Junqi Zhao and Weisheng Wang and Boyang Li and Pascale Fung and Steven Hoi},
      year={2023},
      eprint={2305.06500},
      archivePrefix={arXiv},
      primaryClass={cs.CV},
      url={https://arxiv.org/abs/2305.06500}, 
}

@misc{zhai2024finetuninglargevisionlanguagemodels,
      title={Fine-Tuning Large Vision-Language Models as Decision-Making Agents via Reinforcement Learning}, 
      author={Yuexiang Zhai and Hao Bai and Zipeng Lin and Jiayi Pan and Shengbang Tong and Yifei Zhou and Alane Suhr and Saining Xie and Yann LeCun and Yi Ma and Sergey Levine},
      year={2024},
      eprint={2405.10292},
      archivePrefix={arXiv},
      primaryClass={cs.AI},
      url={https://arxiv.org/abs/2405.10292}, 
}

@misc{wang2024rlvlmfreinforcementlearningvision,
      title={RL-VLM-F: Reinforcement Learning from Vision Language Foundation Model Feedback}, 
      author={Yufei Wang and Zhanyi Sun and Jesse Zhang and Zhou Xian and Erdem Biyik and David Held and Zackory Erickson},
      year={2024},
      eprint={2402.03681},
      archivePrefix={arXiv},
      primaryClass={cs.RO},
      url={https://arxiv.org/abs/2402.03681}, 
}

@misc{ppo,
      title={Proximal Policy Optimization Algorithms}, 
      author={John Schulman and Filip Wolski and Prafulla Dhariwal and Alec Radford and Oleg Klimov},
      year={2017},
      eprint={1707.06347},
      archivePrefix={arXiv},
      primaryClass={cs.LG},
      url={https://arxiv.org/abs/1707.06347}, 
}

@misc{dpo,
      title={Direct Preference Optimization: Your Language Model is Secretly a Reward Model}, 
      author={Rafael Rafailov and Archit Sharma and Eric Mitchell and Stefano Ermon and Christopher D. Manning and Chelsea Finn},
      year={2024},
      eprint={2305.18290},
      archivePrefix={arXiv},
      primaryClass={cs.LG},
      url={https://arxiv.org/abs/2305.18290}, 
}

@article{vlrethinker,
  title={Vl-rethinker: Incentivizing self-reflection of vision-language models with reinforcement learning},
  author={Wang, Haozhe and Qu, Chao and Huang, Zuming and Chu, Wei and Lin, Fangzhen and Chen, Wenhu},
  journal={arXiv preprint arXiv:2504.08837},
  year={2025}
}

@article{lookback,
  title={Look-back: Implicit visual re-focusing in mllm reasoning},
  author={Yang, Shuo and Niu, Yuwei and Liu, Yuyang and Ye, Yang and Lin, Bin and Yuan, Li},
  journal={arXiv preprint arXiv:2507.03019},
  year={2025}
}

@inproceedings{textvqa,
  title={Towards vqa models that can read},
  author={Singh, Amanpreet and Natarajan, Vivek and Shah, Meet and Jiang, Yu and Chen, Xinlei and Batra, Dhruv and Parikh, Devi and Rohrbach, Marcus},
  booktitle={Proceedings of the IEEE/CVF conference on computer vision and pattern recognition},
  pages={8317--8326},
  year={2019}
}

@article{scienceqa,
  title={Learn to explain: Multimodal reasoning via thought chains for science question answering},
  author={Lu, Pan and Mishra, Swaroop and Xia, Tanglin and Qiu, Liang and Chang, Kai-Wei and Zhu, Song-Chun and Tafjord, Oyvind and Clark, Peter and Kalyan, Ashwin},
  journal={Advances in neural information processing systems},
  volume={35},
  pages={2507--2521},
  year={2022}
}

@inproceedings{vqav2,
  title={Making the v in vqa matter: Elevating the role of image understanding in visual question answering},
  author={Goyal, Yash and Khot, Tejas and Summers-Stay, Douglas and Batra, Dhruv and Parikh, Devi},
  booktitle={Proceedings of the IEEE conference on computer vision and pattern recognition},
  pages={6904--6913},
  year={2017}
}

@inproceedings{gqa,
  title={Gqa: A new dataset for real-world visual reasoning and compositional question answering},
  author={Hudson, Drew A and Manning, Christopher D},
  booktitle={Proceedings of the IEEE/CVF conference on computer vision and pattern recognition},
  pages={6700--6709},
  year={2019}
}

@article{gem,
  title={Gradient episodic memory for continual learning},
  author={Lopez-Paz, David and Ranzato, Marc'Aurelio},
  journal={Advances in neural information processing systems},
  volume={30},
  year={2017}
}

@article{mllm,
  title={A survey on multimodal large language models},
  author={Yin, Shukang and Fu, Chaoyou and Zhao, Sirui and Li, Ke and Sun, Xing and Xu, Tong and Chen, Enhong},
  journal={National Science Review},
  volume={11},
  number={12},
  pages={nwae403},
  year={2024},
  publisher={Oxford University Press}
}

@misc{perceptionr1,
      title={Perception-R1: Pioneering Perception Policy with Reinforcement Learning}, 
      author={En Yu and Kangheng Lin and Liang Zhao and Jisheng Yin and Yana Wei and Yuang Peng and Haoran Wei and Jianjian Sun and Chunrui Han and Zheng Ge and Xiangyu Zhang and Daxin Jiang and Jingyu Wang and Wenbing Tao},
      year={2025},
      eprint={2504.07954},
      archivePrefix={arXiv},
      primaryClass={cs.CV},
      url={https://arxiv.org/abs/2504.07954}, 
}

@article{deepseekr1,
   title={DeepSeek-R1 incentivizes reasoning in LLMs through reinforcement learning},
   volume={645},
   ISSN={1476-4687},
   url={http://dx.doi.org/10.1038/s41586-025-09422-z},
   DOI={10.1038/s41586-025-09422-z},
   number={8081},
   journal={Nature},
   publisher={Springer Science and Business Media LLC},
   author={Guo, Daya and Yang, Dejian and Zhang, Haowei and Song, Junxiao and Wang, Peiyi and Zhu, Qihao and Xu, Runxin and Zhang, Ruoyu and Ma, Shirong and Bi, Xiao and Zhang, Xiaokang and Yu, Xingkai and Wu, Yu and Wu, Z. F. and Gou, Zhibin and Shao, Zhihong and Li, Zhuoshu and Gao, Ziyi and Liu, Aixin and Xue, Bing and Wang, Bingxuan and Wu, Bochao and Feng, Bei and Lu, Chengda and Zhao, Chenggang and Deng, Chengqi and Ruan, Chong and Dai, Damai and Chen, Deli and Ji, Dongjie and Li, Erhang and Lin, Fangyun and Dai, Fucong and Luo, Fuli and Hao, Guangbo and Chen, Guanting and Li, Guowei and Zhang, H. and Xu, Hanwei and Ding, Honghui and Gao, Huazuo and Qu, Hui and Li, Hui and Guo, Jianzhong and Li, Jiashi and Chen, Jingchang and Yuan, Jingyang and Tu, Jinhao and Qiu, Junjie and Li, Junlong and Cai, J. L. and Ni, Jiaqi and Liang, Jian and Chen, Jin and Dong, Kai and Hu, Kai and You, Kaichao and Gao, Kaige and Guan, Kang and Huang, Kexin and Yu, Kuai and Wang, Lean and Zhang, Lecong and Zhao, Liang and Wang, Litong and Zhang, Liyue and Xu, Lei and Xia, Leyi and Zhang, Mingchuan and Zhang, Minghua and Tang, Minghui and Zhou, Mingxu and Li, Meng and Wang, Miaojun and Li, Mingming and Tian, Ning and Huang, Panpan and Zhang, Peng and Wang, Qiancheng and Chen, Qinyu and Du, Qiushi and Ge, Ruiqi and Zhang, Ruisong and Pan, Ruizhe and Wang, Runji and Chen, R. J. and Jin, R. L. and Chen, Ruyi and Lu, Shanghao and Zhou, Shangyan and Chen, Shanhuang and Ye, Shengfeng and Wang, Shiyu and Yu, Shuiping and Zhou, Shunfeng and Pan, Shuting and Li, S. S. and Zhou, Shuang and Wu, Shaoqing and Yun, Tao and Pei, Tian and Sun, Tianyu and Wang, T. and Zeng, Wangding and Liu, Wen and Liang, Wenfeng and Gao, Wenjun and Yu, Wenqin and Zhang, Wentao and Xiao, W. L. and An, Wei and Liu, Xiaodong and Wang, Xiaohan and Chen, Xiaokang and Nie, Xiaotao and Cheng, Xin and Liu, Xin and Xie, Xin and Liu, Xingchao and Yang, Xinyu and Li, Xinyuan and Su, Xuecheng and Lin, Xuheng and Li, X. Q. and Jin, Xiangyue and Shen, Xiaojin and Chen, Xiaosha and Sun, Xiaowen and Wang, Xiaoxiang and Song, Xinnan and Zhou, Xinyi and Wang, Xianzu and Shan, Xinxia and Li, Y. K. and Wang, Y. Q. and Wei, Y. X. and Zhang, Yang and Xu, Yanhong and Li, Yao and Zhao, Yao and Sun, Yaofeng and Wang, Yaohui and Yu, Yi and Zhang, Yichao and Shi, Yifan and Xiong, Yiliang and He, Ying and Piao, Yishi and Wang, Yisong and Tan, Yixuan and Ma, Yiyang and Liu, Yiyuan and Guo, Yongqiang and Ou, Yuan and Wang, Yuduan and Gong, Yue and Zou, Yuheng and He, Yujia and Xiong, Yunfan and Luo, Yuxiang and You, Yuxiang and Liu, Yuxuan and Zhou, Yuyang and Zhu, Y. X. and Huang, Yanping and Li, Yaohui and Zheng, Yi and Zhu, Yuchen and Ma, Yunxian and Tang, Ying and Zha, Yukun and Yan, Yuting and Ren, Z. Z. and Ren, Zehui and Sha, Zhangli and Fu, Zhe and Xu, Zhean and Xie, Zhenda and Zhang, Zhengyan and Hao, Zhewen and Ma, Zhicheng and Yan, Zhigang and Wu, Zhiyu and Gu, Zihui and Zhu, Zijia and Liu, Zijun and Li, Zilin and Xie, Ziwei and Song, Ziyang and Pan, Zizheng and Huang, Zhen and Xu, Zhipeng and Zhang, Zhongyu and Zhang, Zhen},
   year={2025},
   month=Sept, pages={633–638} }


\appendix

\newpage
\section{Appendix}

\subsection{Symbols}

We use the following notation throughout the paper:
\begin{itemize}
    \item $\mathcal{T}$: a sequence task.
    \item $t$: total number of tasks in the continual learning sequence.
    \item $T_i$: dataset of task $i$; a sample is a visual–question pair $x_i\!\sim\!T_i$.
    \item $\pi_\theta$: current MLLM with parameters $\theta$; $\pi_{\theta_i}$ denotes the model after completing task $i$.
    \item $\pi_{\mathrm{ref}}$: frozen reference model, set to $\pi_{\theta_{i-1}}$ at the start of task $i$.
    \item $r$, $a$: reasoning trajectory and its final answer token(s), respectively.
          We denote $(r^{i-1},a^{i-1})\!\sim\!\pi_{\theta_{i-1}}(\cdot\!\mid\!x)$ for samples from the previous model and
          $(r^{i},a^{i})\!\sim\!\pi_{\theta_i}(\cdot\!\mid\!x)$ from the current one.
    \item $C_{\theta}(\cdot\!\mid\!x)$: confidence score for outputs under parameters $\theta$.
    \item $\tau$:  the threshold for reasoning score to evaluate its reasoning portability.
\end{itemize}

\subsection{More Results of Experiments}
  
\paragraph{Further Reasoning Confidence Reliability Analysis.}

\begin{figure}[t]
  \centering
  
  \begin{minipage}[t]{0.45\linewidth}
  \vspace{0pt}
  \centering
    \includegraphics[width=1\linewidth]{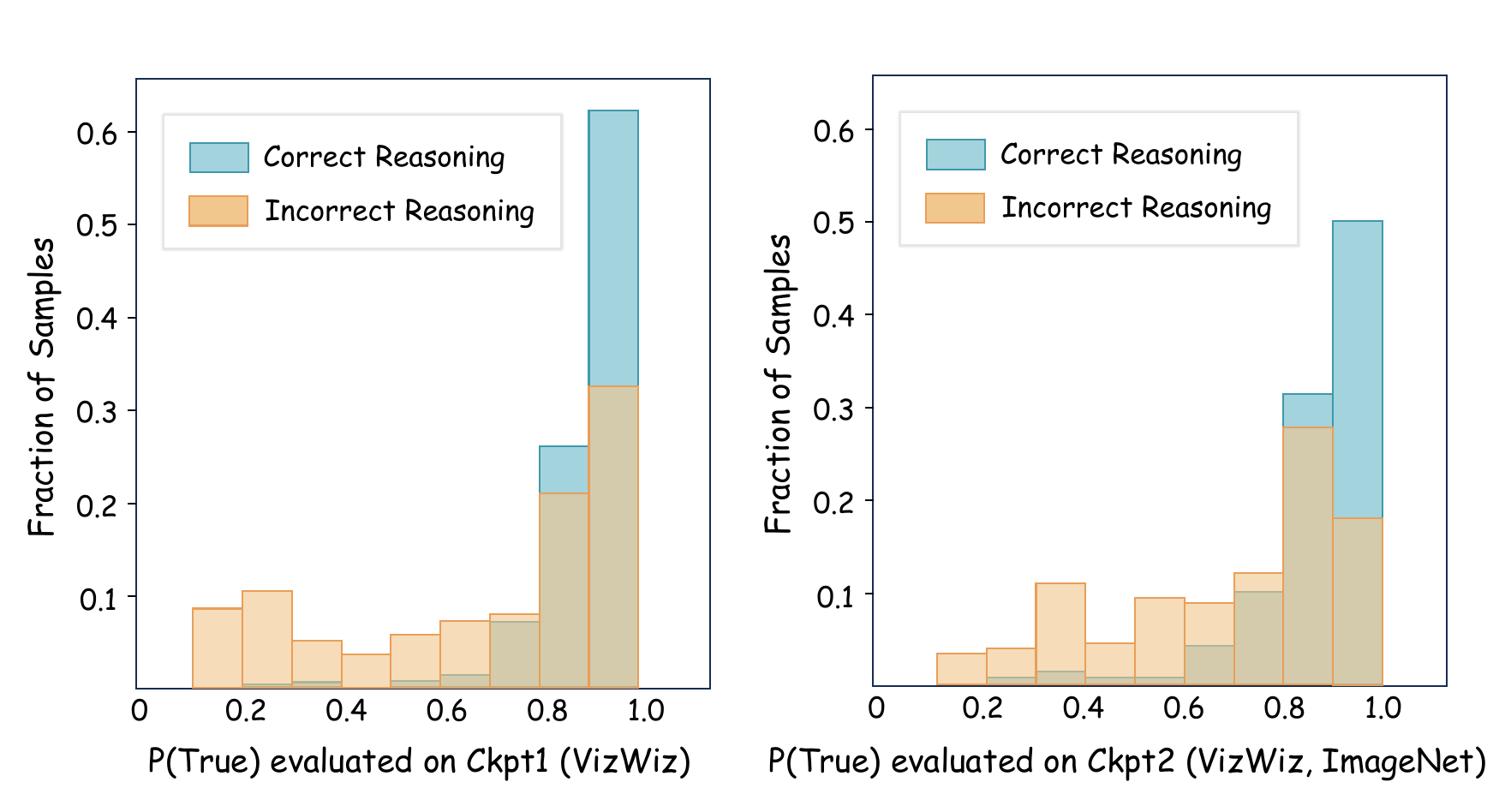}
  \caption{Reasoning confidence distribution on IconQA. \textit{Left}: ckpt1 (after VizWiz). \textit{Right}: ckpt2 (after VizWiz$\rightarrow$ImageNet).}
  \label{fig:rc_degradation}
  \end{minipage}
  \hfill
   \begin{minipage}[t]{0.53\linewidth}
  \vspace{0pt}
    \centering
    \includegraphics[width=\linewidth]{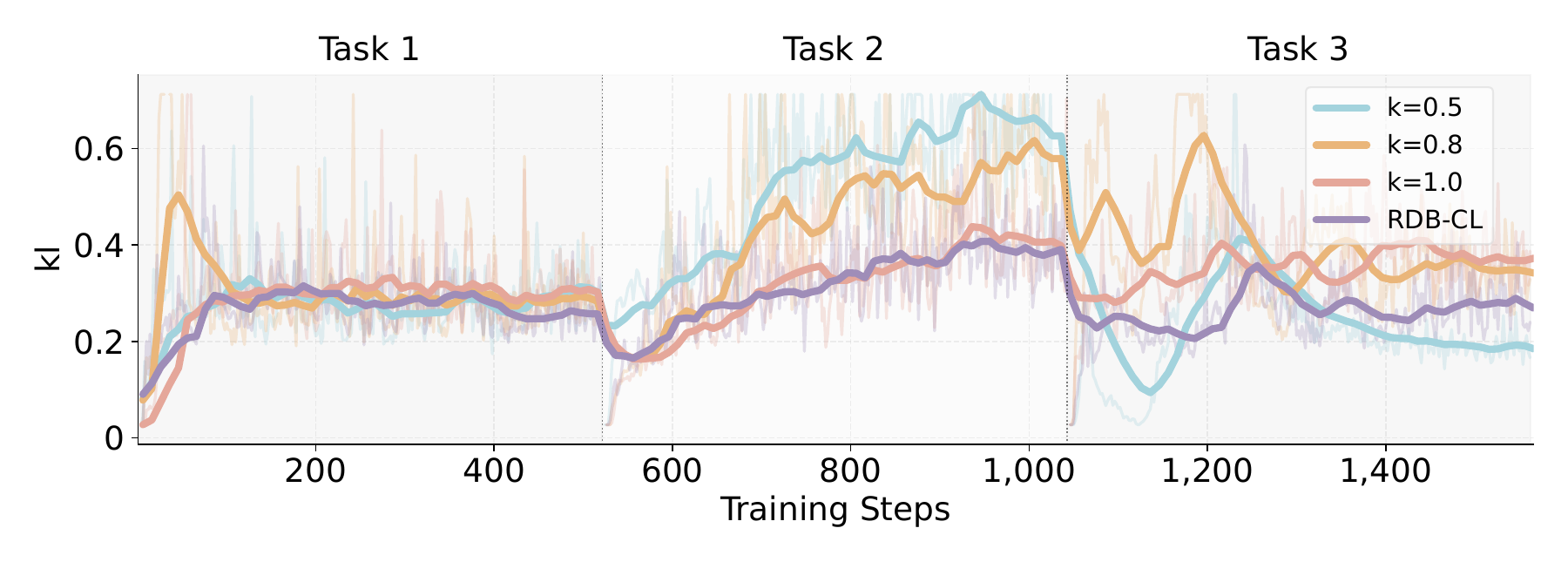}
    \captionof{figure}{Training dynamics of KL divergence under different KL weights, which is regarded as a strong predictor of catastrophic forgetting~\cite{RLRazor}. RDB-CL exhibits the most stable KL curve, demonstrating robustness against distribution shift.}
    \label{fig:kl_dynamics}
  \end{minipage}
\end{figure}

To examine whether the reliability of reasoning confidence degrades as the model acquires more tasks in the continual learning context, we further evaluate the distribution for IconQA on two sequential checkpoints. As shown in \cref{fig:rc_degradation}, reasoning confidence both preserves a clear separability between correct and incorrect reasoning. Notably, the distinction in the low-confidence region remains highly stable, indicating strong robustness against task interference. It suggests that reasoning confidence serves as a reliable indicator throughout continual learning, capturing a deeper signal of reasoning portability across tasks, rather than stemming from the zero-shot capability of MLLMs.

\begin{figure}[h]
  \centering
  \includegraphics[height=3.4cm]{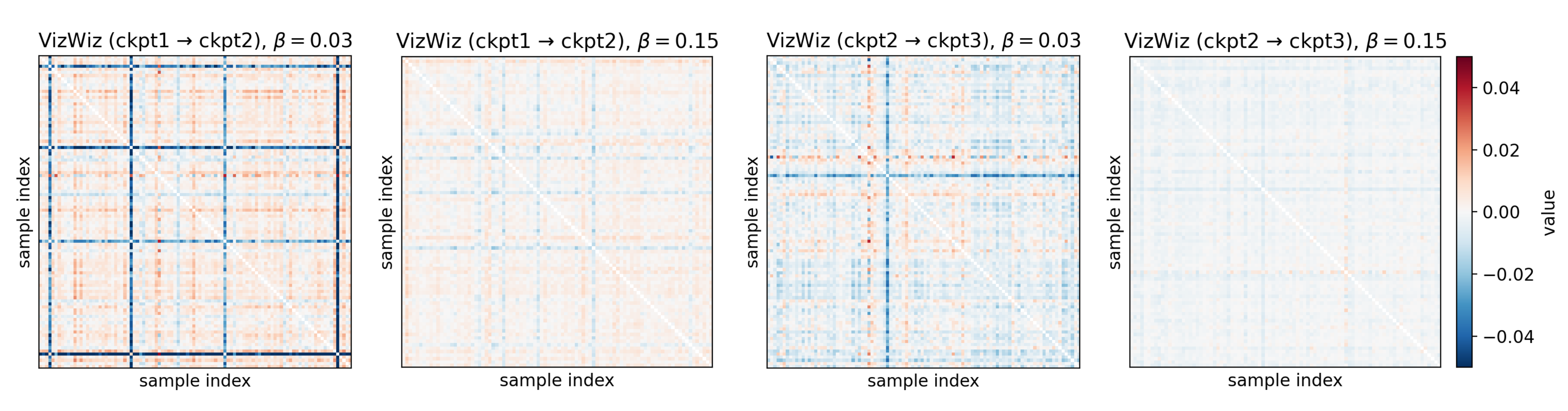}
  \caption{Heatmap of pairwise feature-distance changes among VizWiz samples across sequentially trained models with different KL coefficient $\beta$. \emph{From left to right}: (1) \emph{ckpt1 $\rightarrow$ ckpt2 }(after VizWiz $\rightarrow$ after ImageNet), \emph{$\beta= 0.03$}; (2) \emph{ckpt1 $\rightarrow$ ckpt2 }, \emph{$\beta= 0.15$}; (3) \emph{ckpt2 $\rightarrow$ ckpt3 } (after ImageNet $\rightarrow$ after IconQA), \emph{$\beta= 0.03$}; (4) \emph{ckpt2 $\rightarrow$ ckpt3 }, \emph{$\beta= 0.15$}. All models are trained with GRPO. A stronger KL constraint suppresses feature drift, whereas a weaker one allows greater adaptability.
  }
  \label{fig:feature_shift}
\end{figure}

\paragraph{Empirical Study of Static KL in Continual Learning.}
We provide an additional analysis of how static KL regularization affects representation drift in continual learning. Specifically, we measure the feature shift under different fixed KL coefficients $\beta$ as shown in \cref{fig:feature_shift}. This observation further motivates our portability-aware dynamic KL modulation.

\vspace{2em}
\paragraph{Generalization to DAPO.}
To further demonstrate the generality and robustness of our approach, we extend RDB-CL to DAPO~\cite{dapo}, another representative RLVR algorithm that adopts a KL-regularized variant. As reported in \cref{tab:dapo_result}, RDB-CL delivers consistent performance gains on top of DAPO, indicating that its benefits are not tied to GRPO but generalize across different RLVR frameworks.
\vspace{2em}
\newcolumntype{C}[1]{>{\centering\arraybackslash}p{#1}}

\begin{center}
\small
\captionof{table}{Results on DAPO and RDB-CL based DAPO.}
\label{tab:dapo_result}
\setlength{\tabcolsep}{6pt}
\renewcommand{\arraystretch}{1.2}
\begin{tabular}{p{2.4cm} C{1.3cm} C{1.3cm} C{1.3cm} C{1.3cm}}
\toprule
Method & \textit{Avg.} & \textit{Last} & \textit{Finetune} & \textit{BwT} \\
\midrule
DAPO
& \numplus{46.85}{(+0.0\%)}
& \numplus{55.93}{(+0.0\%)}
& \numplus{58.85}{(+0.0\%)}
& \numplus{-4.37}{(+0.0\%)} \\

\rowcolor{gray!10}
RDB-CL
& \numplustop{\textbf{47.65}}{(+1.7\%)}
& \numplustop{\textbf{57.66}}{(+3.1\%)}
& \numplustop{\textbf{59.48}}{(+1.1\%)}
& \numplustop{\textbf{-2.72}}{(+37.8\%)} \\
\bottomrule
\end{tabular}
\end{center}

\paragraph{Detailed Results.}
In this subsection, we present the full per-task results across different streaming checkpoints for the main experiment. Columns correspond to checkpoints obtained after sequentially training on each task (ckpt1: VizWiz → ckpt2: ImageNet → ckpt3: IconQA), while rows report the performance for each task.
\newcolumntype{C}[1]{>{\centering\arraybackslash}p{#1}}
\captionsetup[table]{skip=2pt}

\begin{center}
\small
\captionof{table}{Detailed results under GRPO.}
\setlength{\tabcolsep}{6pt}
\renewcommand{\arraystretch}{1.2}
\begin{tabular}{p{2.4cm} C{1.3cm} C{1.3cm} C{1.3cm} C{1.3cm}}
\toprule
Task     & ckpt1 & ckpt2 & ckpt3 & Average \\
\midrule
VizWiz   & 53.75 & 49.93 & 49.57 & 51.08 \\
ImageNet & 19.76 & 62.12 & 50.63 & 44.17 \\
IconQA   & 31.77 & 18.91 & 36.01 & 28.90 \\
\bottomrule
\end{tabular}
\end{center}

\vspace{2pt}

\begin{center}
\small
\captionof{table}{Detailed results under RDB-CL.}
\setlength{\tabcolsep}{6pt}
\renewcommand{\arraystretch}{1.2}
\begin{tabular}{p{2.4cm} C{1.3cm} C{1.3cm} C{1.3cm} C{1.0cm}}
\toprule
Task     & ckpt1 & ckpt2 & ckpt3 & Average \\
\midrule
VizWiz   & 54.58 & 52.40 & 50.54 & 52.51 \\
ImageNet & 18.81 & 63.96 & 58.34 & 47.04 \\
IconQA   & 32.53 & 16.79 & 52.50 & 33.94 \\
\bottomrule
\end{tabular}
\end{center}

\vspace{2pt}

\begin{center}
\small
\captionof{table}{Detailed results for Qwen2.5-VL-7B-Instruct.}
\setlength{\tabcolsep}{6pt}
\renewcommand{\arraystretch}{1.2}
\begin{tabular}{p{2.4cm} C{1.3cm} C{1.3cm} C{1.3cm} C{1.3cm}}
\toprule
           & VizWiz & ImageNet & IconQA & Average \\
\midrule
Zero-shot  & 40.47  & 19.37    & 39.64  & 33.16 \\
\bottomrule
\end{tabular}
\end{center}

\vspace{2pt}

\begin{center}
\small
\captionof{table}{Detailed results under SFT (Multi-tasks).}
\setlength{\tabcolsep}{6pt}
\renewcommand{\arraystretch}{1.2}
\begin{tabular}{p{2.4cm} C{1.3cm} C{1.3cm} C{1.3cm} C{1.3cm}}
\toprule
              & VizWiz & ImageNet & IconQA & Average \\
\midrule
Multi-tasks   & 22.95  & 75.21    & 66.68  & 54.95 \\
\bottomrule
\end{tabular}
\end{center}

\vspace{2pt}

\begin{center}
\small
\captionof{table}{Detailed results under SFT (In-order).}
\setlength{\tabcolsep}{6pt}
\renewcommand{\arraystretch}{1.2}
\begin{tabular}{p{2.4cm} C{1.3cm} C{1.3cm} C{1.3cm} C{1.3cm}}
\toprule
Task     & ckpt1 & ckpt2 & ckpt3 & Average \\
\midrule
VizWiz   & 45.61 & 25.03 & 12.77 & 27.80 \\
ImageNet & 17.07 & 83.59 & 72.49 & 57.72 \\
IconQA   & 31.92 & 33.13 & 63.99 & 43.01 \\
\bottomrule
\end{tabular}
\end{center}

\vspace{2pt}

\begin{center}
\small
\captionof{table}{Detailed results under GRPO training with EWC.}
\setlength{\tabcolsep}{6pt}
\renewcommand{\arraystretch}{1.2}
\begin{tabular}{p{2.4cm} C{1.3cm} C{1.3cm} C{1.3cm} C{1.3cm}}
\toprule
Task     & ckpt1 & ckpt2 & ckpt3 & Average \\
\midrule
VizWiz   & 53.23 & 50.37 & 48.13 & 50.58 \\
ImageNet & 19.33 & 61.62 & 56.73 & 45.89 \\
IconQA   & 23.30 & 15.13 & 49.62 & 29.35 \\
\bottomrule
\end{tabular}
\end{center}

\vspace{2pt}

\begin{center}
\small
\captionof{table}{Detailed results under GRPO training with LwF.}
\setlength{\tabcolsep}{6pt}
\renewcommand{\arraystretch}{1.2}
\begin{tabular}{p{2.4cm} C{1.3cm} C{1.3cm} C{1.3cm} C{1.3cm}}
\toprule
Task     & ckpt1 & ckpt2 & ckpt3 & Average \\
\midrule
VizWiz   & 51.48 & 51.04 & 50.04 & 50.85 \\
ImageNet & 19.33 & 47.31 & 46.10 & 37.58 \\
IconQA   & 33.59 & 20.59 & 62.90 & 39.03 \\
\bottomrule
\end{tabular}
\end{center}
\vspace{2pt}
\begin{center}
\small
\captionof{table}{Detailed results under MoELoRA.}
\setlength{\tabcolsep}{6pt}
\renewcommand{\arraystretch}{1.2}
\begin{tabular}{p{2.4cm} C{1.3cm} C{1.3cm} C{1.3cm} C{1.3cm}}
\toprule
Task     & ckpt1 & ckpt2 & ckpt3 & Average \\
\midrule
VizWiz   & 31.90 & 30.24 & 32.37 & 31.50 \\
ImageNet & 18.02 & 21.55 & 19.71 & 19.76 \\
IconQA   & 46.14 & 45.99 & 70.35 & 54.16 \\
\bottomrule
\end{tabular}
\end{center}
\vspace{2pt}
\begin{center}
\small
\captionof{table}{Detailed results under HiDe-LLaVA.}
\setlength{\tabcolsep}{6pt}
\renewcommand{\arraystretch}{1.2}
\begin{tabular}{p{2.4cm} C{1.3cm} C{1.3cm} C{1.3cm} C{1.3cm}}
\toprule
Task     & ckpt1 & ckpt2 & ckpt3 & Average \\
\midrule
VizWiz   & 27.03 & 26.17 & 28.52 & 27.24 \\
ImageNet & 17.35 & 27.13 & 26.74 & 23.74 \\
IconQA   & 48.11 & 47.96 & 71.26 & 55.78 \\
\bottomrule
\end{tabular}
\end{center}
\label{app:more_results}

\subsection{Cost Analysis}
\label{subsec:cost}
RDB-CL introduces \textit{no additional trainable parameters} and requires \textit{no extra memory} beyond standard RLVR training. It also incurs \textit{no inference-time latency}, since reasoning confidence is only computed during training.

\paragraph{Training overhead.}
RDB-CL estimates reasoning confidence via a single-token P(True) probe, which adds only marginal compute per sample. In our implementation, the overall wall-clock training time increases by approximately $\sim$13.6\%, primarily due to extra forward passes when the probe is not fully fused with rollout generation. Despite this overhead, the additional compute is modest relative to the performance benefits (e.g., \textbf{+12.0\%} \textit{Last} accuracy in the main benchmark).
In contrast, replay-based methods often substantially increase training time, while Mixture-of-Experts (MoE)-style approaches can significantly enlarge memory footprint due to additional expert parameters. RDB-CL achieves strong continual learning gains without introducing such memory or time costs.

\subsection{More Training Details}
\label{app:train_details}
In addition to the main setup described in \cref{subsec:exp_setup}, we provide here the remaining implementation details for completeness, following the reference implementations~\cite{coin, tan2025reason}.
Our reward function consists of an \textit{accuracy reward} and a \textit{format reward}. The \textit{accuracy reward} evaluates only the final answer, assigning $1.0$ for correct predictions and $0$ otherwise. The \textit{format reward} is based on strict matching of the required \texttt{<think>...</think><answer>...</answer>} structure, assigning $0.7$ for perfect formatting, $0$ for partially correct formatting, and $-0.7$ when the constraint is violated.
All methods share the same optimization hyperparameters:  an optimizer AdamW with $\beta_2 = 0.95$, weight decay of 0.1, a cosine learning-rate schedule with a warm-up ratio of $0.01$, batch size of $1$ per device with gradient accumulation of $2$, and training for one epoch. The rollout number for reference model is $1$. Rollouts are sampled with temperature $0.7$, top-$p=0.9$, top-$k=50$, a maximum generation length of $8192$ tokens, and image preprocessing that isotropically rescales inputs such that their pixel count lies within $[3136, 440000]$.
Besides, following the GRPO implementation we build upon~\cite{tan2025reason}, wherein $\pi_{\theta_{\mathrm{old}}}(\cdot)=\pi_{\theta}(\cdot)$ and clipping on the advantage term is therefore omitted, we retain this setting to isolate the effect of KL regularization in our analysis.
We also plot the evolution of the KL divergence during training in \cref{fig:kl_dynamics}. 

\begin{figure*}[h]
  \centering
    \includegraphics[width=1\linewidth]{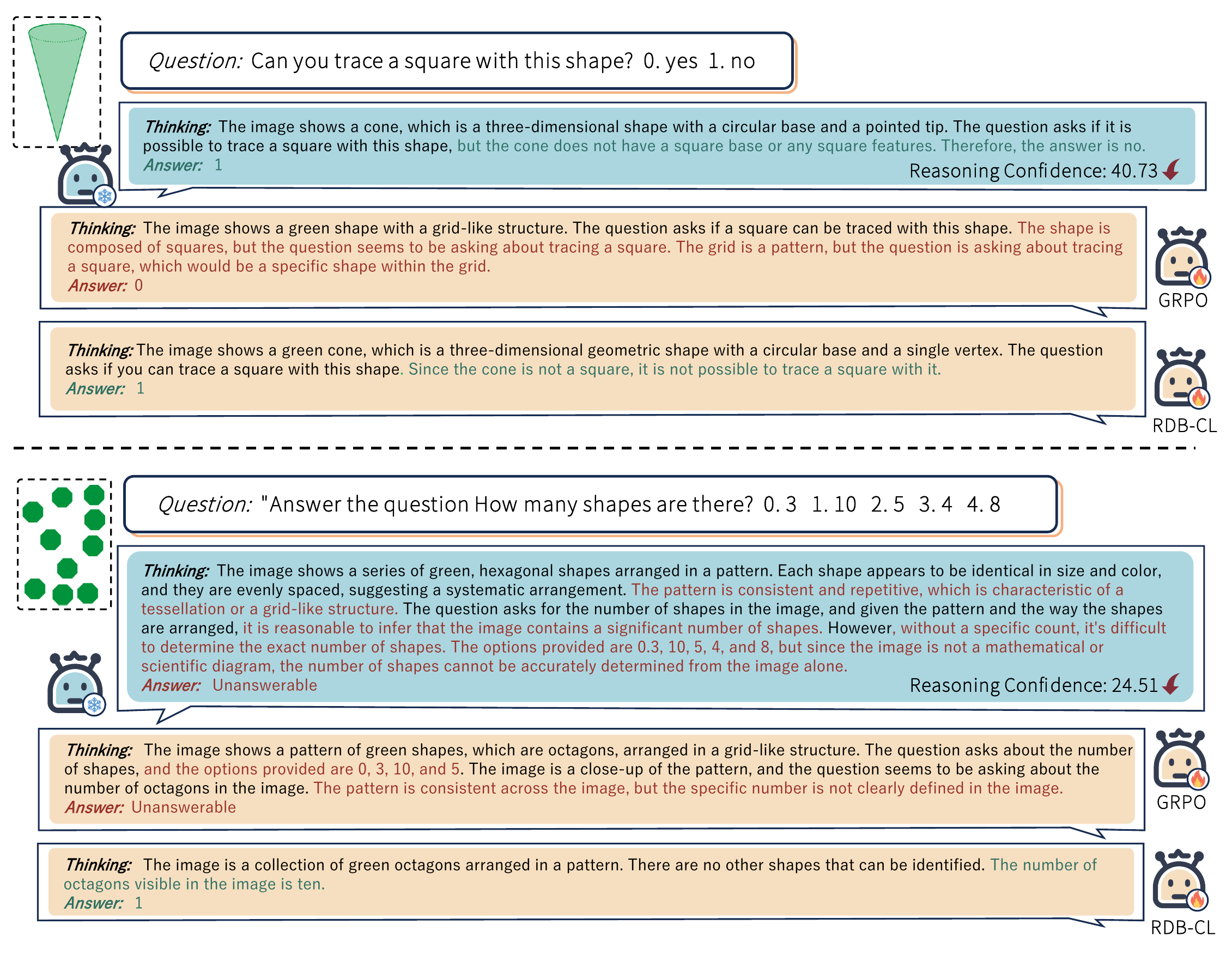}
  \caption{\textit{Qualitative comparison under representative low reasoning confidence scenarios.} Low-confidence cases typically arise from two sources: domain-gap yet \textcolor[HTML]{2E7064}{sound} reasoning(\textit{Top}) and \textcolor[HTML]{9B3128}{incoherent} reasoning(\textit{Bottom}). After training, RDB-CL generates more \textcolor[HTML]{2E7064}{concise and reliable} reasoning traces, while GRPO tends to produce \textcolor[HTML]{9B3128}{incorrect or irrelevant} text.}
  \label{fig:sample_low}
\end{figure*}

\subsection{Qualitative Comparisons under Broader Reasoning Confidence Scenarios}
\label{subsec:more_confidence_cases}

\begin{figure*}[h]
  \centering
    \includegraphics[width=1\linewidth]{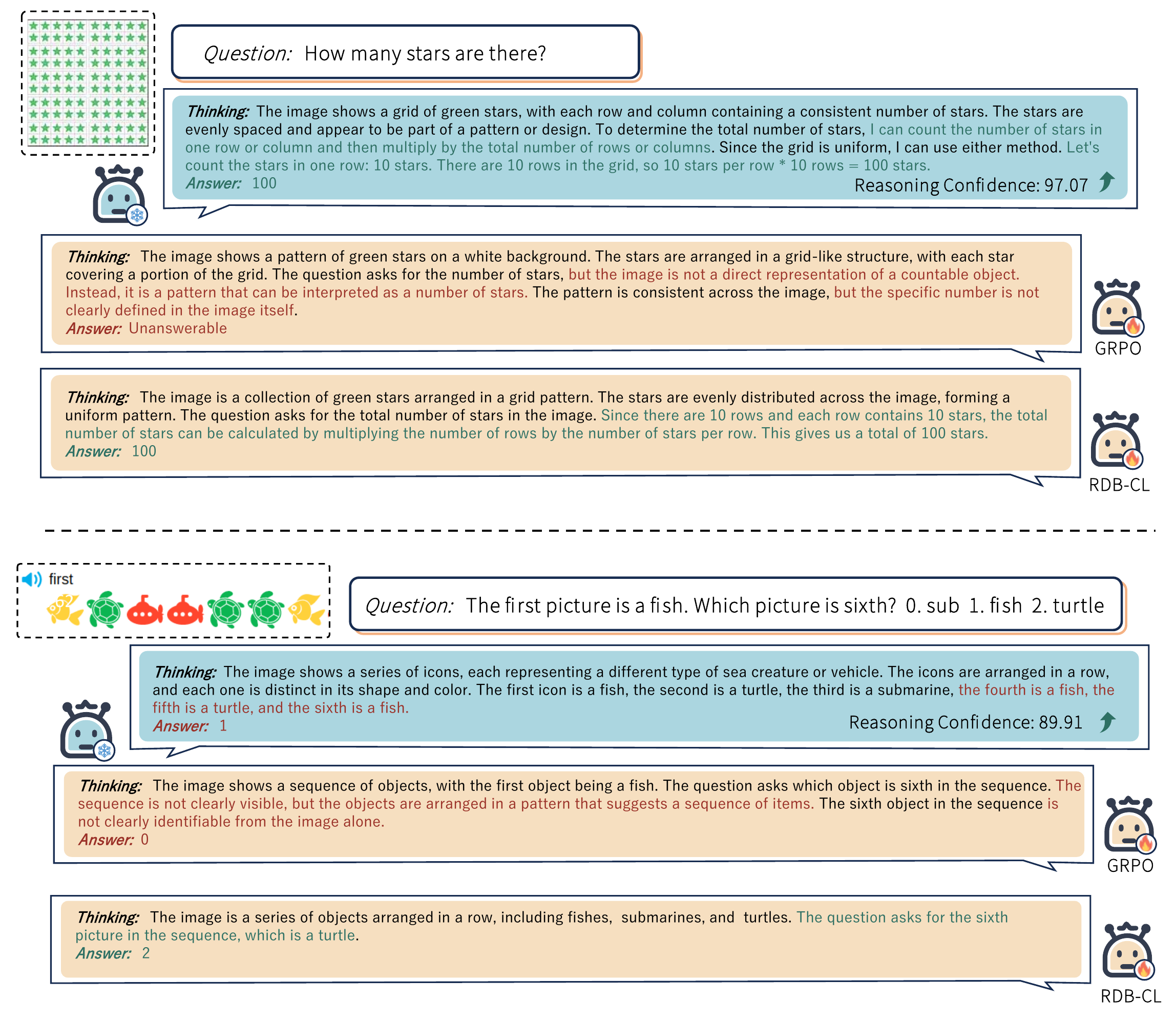}
  \caption{\textit{Qualitative comparison under representative high reasoning confidence scenarios.} High-confidence cases typically fall into two types: (\textit{Top}) cases where the policy successfully produces task-solving reasoning for new tasks, and (\textit{Bottom}) cases with seemingly sound reasoning that still fail due to missing task-specific, non-reasoning knowledge.}
  \label{fig:sample_high}
\end{figure*}

Beyond the low-confidence case study in \cref{subsec:confidence_case}, we provide additional qualitative comparisons across broader reasoning confidence scenarios. \cref{fig:sample_low} presents representative low-confidence cases, while \cref{fig:sample_high} shows high-confidence cases.
We further find that questions of the same type tend to yield similar reasoning confidence, reflecting their underlying logical consistency. Comparing the outputs of GRPO and RDB-CL across these scenarios reveals several notable differences. GRPO tends to consistently follow patterns inherited from previously learned tasks, as evidenced by the similarity in its output formats. However, due to its over-constrained behavior and limited exploration, it can deviate even when the reference model already encodes the correct reasoning trajectory. In such cases, the compressed task space and interference from earlier tasks cause GRPO to overwrite previously correct capabilities, leading to catastrophic forgetting (as illustrated in \cref{fig:sample_high} \textit{Top}). 
In contrast, RDB-CL learns task-specific patterns for new tasks. By leveraging the reasoning confidence score to modulate learning behaviors dynamically, it explores proactively under low confidence while preserving previously acquired reasoning abilities under high confidence. This adaptive mechanism enables more stable generalization and effective continual reasoning.

\subsection{Definitions and Preliminaries }
\label{app:port_define}
\paragraph{Definition (Portability).}
Given a sample $x$ from the new task $T_t$ and the previous policy $\pi_{\theta_{t-1}}$, portability $P(x; \pi_{\theta_{t-1}}) \in [0,1]$ measures the extent to which $\pi_{\theta_{t-1}}$'s behavior on $x$ can be reused without revision. We distinguish two regimes:
\begin{itemize}
    \item \textbf{High Portability} : the previous policy's behavior on $x$ is applicable, retaining cross-task generalization value.
    \item \textbf{Low Portability} : the previous policy's behavior is misaligned with $x$, indicating a need for adaptation to acquire new capability.
\end{itemize}

\paragraph{Preliminary: Reasoning via RLVR} ~\label{subsec:back}
Unlike RLHF~\cite{ppo}, RLVR~\cite{tulu3} provides scalable and reliable supervision for reasoning tasks by using task-grounded rewards. Given a MLLM parameterized by $\theta$, a dataset $D$, and a reward verifier $R_{\text{ver}}$, the objective of RLVR is:
\begin{equation}
    \max_{\theta} \mathbb{E}_{x \sim D, y \sim \pi_{\theta}(\cdot|x)} [R_{ver}(x, y)]
    \label{eq:RLVRobjective},
\end{equation}
where $x$ is the input and $y$ is the output produced by the policy $\pi_{\theta}(y|x)$. Our work builds on GRPO~\cite{deepseekmath}, a representative RLVR algorithm. GRPO updates the policy using group-relative advantages and a KL-regularized objective. Its loss function is:
\begin{equation}
\begin{split}
\mathcal{J}_{GRPO}(\theta) = - \frac{1}{G} \sum_{i=1}^G \frac{1}{|o_i|} 
\sum_{t=1}^{|o_i|} \left[ r_\theta(q,o_i,t) A_{i,t} \right. 
\left. - \beta D_{\text{KL}} (\pi_{\theta} || \pi_{\text{ref}}) \right],
\end{split}
 \label{eq:GRPOloss}
\end{equation}
where $A_{i,t}$ is the advantage, $r_{\theta}$ is the likelihood ratio, and the $D_{\text{KL}}$ term constrains the policy from deviating from a reference policy $\pi_{\text{ref}}$, whose strength is controlled by a coefficient $\beta$.

\subsection{Prompts}

\tcbset{
  mybox/.style={
    enhanced,            
    breakable,           
    colback=gray!10,     
    colframe=black!70,   
    arc=1mm,             
    boxrule=0.6pt,       
    left=6mm,right=6mm,
    top=4mm,bottom=4mm,
    colbacktitle=black,  
    coltitle=white,      
    fonttitle=\bfseries, 
    boxed title style={
      size=small,        
      arc=0mm, outer arc=0mm, 
      boxrule=0pt        
    },
  }
}

\paragraph{Training and Evaluation.}
We adopt an identical system prompt and user prompt for all tasks during both training and evaluation. 

\begin{tcolorbox}[enhanced, colback=gray!2, colframe=black, fonttitle=\bfseries,
                  title=SYSTEM\_PROMPT, sharp corners=south, boxrule=0.8pt, 
                  colbacktitle=black, coltitle=white, left=6pt, right=6pt, top=6pt, bottom=6pt]

\textbf{A conversation between User and Assistant.}  
The User asks a question, and the Assistant solves it.  
The Assistant first thinks about the reasoning process in the mind and then provides the User with the answer in a word or a phrase as required.

\medskip
You should output the reasoning process first, and then output the answer.  
The reasoning process and answer are enclosed within \texttt{<think>} and \texttt{<answer>} tags, respectively. 

\textbf{Format example:}

{\small\noindent\ttfamily
<think>reasoning process here</think>

<answer>answer here</answer>}

\end{tcolorbox}

\begin{tcolorbox}[enhanced, colback=gray!2, colframe=black, fonttitle=\bfseries,
                  title=USER\_PROMPT, sharp corners=south, boxrule=0.8pt, 
                  colbacktitle=black, coltitle=white, left=6pt, right=6pt, top=6pt, bottom=6pt]
\textbf{**Question**}: \{question\}

\medskip
**question**. First do reasoning and give the answer as required. 
\medskip

Output the \textbf{thinking process} in \texttt{<think> </think>} 

and \textbf{final answer} in \texttt{<answer></answer>} tags.

\end{tcolorbox}

\paragraph{Reasoning Confidence Estimation.}
Following \cite{ptrue}, we use the following prompts to evaluate the model's self-confidence for reasoning.

\begin{tcolorbox}[enhanced, colback=gray!2, colframe=black, fonttitle=\bfseries,
                  title=SYSTEM\_CONFIDENCE\_
                    PROMPT, sharp corners=south, boxrule=0.8pt, 
                  colbacktitle=black, coltitle=white, left=6pt, right=6pt, top=6pt, bottom=6pt]

\textbf{You are an assistant for estimating the reasoning confidence}. The User gives an image with a question, and his own reasoning process, you should evaluate the reasoning is correct, giving the corresponding option directly without any extra text.

\end{tcolorbox}

\begin{tcolorbox}[enhanced, colback=gray!2, colframe=black, fonttitle=\bfseries,
                  title=USER\_CONFIDENCE\_PROMPT, sharp corners=south, boxrule=0.8pt, 
                  colbacktitle=black, coltitle=white, left=6pt, right=6pt, top=6pt, bottom=6pt]
Answer whether the **reasoning** is correct for the given **question**. 
\medskip

\textbf{**Question**}: \{question\}

\textbf{**Reasoning**}: \{reasoning\}

\medskip
Is the **reasoning** correct?

A) True
   B) False

\medskip
The **reasoning** is: [A / B], depending whether the **reasoning** is correct given the **question** with the image.

Only output A or B in one letter, without any extra text.

\end{tcolorbox}

\subsection{Broader impacts}
\label{subsec:broader_impacts}
This work, to the best of our knowledge, is the first to reveal the role of reasoning confidence in continual learning for vision-language models, and to introduce the concept of \emph{reasoning portability} as a guiding principle for continual adaptation. By moving beyond parameter- or representation-level signals and instead leveraging signals at the reasoning level, our work opens up a new perspective on how continual learning can be informed by what models actually reason about, rather than merely how their weights or features evolve. We hope this perspective will inspire the community to explore reasoning portability, as well as a broader family of reasoning-level signals, for guiding continual learning in realistic, large-scale deployment scenarios where vision-language models must continuously absorb new knowledge without compromising prior capabilities. More generally, we believe this direction can stimulate further research at the intersection of reasoning analysis and continual learning, encouraging future methods that treat reasoning behavior as a first-class citizen in the design of lifelong learning systems.


\end{document}